\documentclass[10pt,twocolumn,letterpaper]{article}

\usepackage{iccv}
\usepackage{times}
\usepackage{epsfig}
\usepackage{graphicx}
\usepackage{amsmath}
\usepackage{amssymb}
\usepackage{algorithm}
\usepackage{multirow}
\usepackage{authblk}

\newcommand*\samethanks[1][\value{footnote}]{\footnotemark[#1]}
\usepackage{xcolor}
\usepackage{algpseudocode}
\usepackage{caption}
\usepackage{subfigure}
\usepackage[accsupp]{axessibility}  
\usepackage[title]{appendix}


\usepackage[breaklinks=true,bookmarks=false]{hyperref}

\iccvfinalcopy 

\def\iccvPaperID{3616} 

\ificcvfinal\fi

\begin{document}

\vspace{-0.5cm}
\title{Influence Selection for Active Learning}
\vspace{-0.5cm}
\author[1]{Zhuoming Liu\thanks{This work was done during their internship at SenseTime Research.}\thanks{Equal contribution.}}
\author[2]{Hao Ding\samethanks[1]\samethanks}
\author[3]{Huaping Zhong}
\author[3,4]{Weijia Li\thanks{Corresponding author}}
\author[3]{Jifeng Dai}
\author[3]{Conghui He}
\affil[1]{University Southern California}
\affil[2]{Johns Hopkins University} 
\affil[3]{SenseTime Research}
\affil[4]{CUHK-SenseTime Joint Lab, The Chinese University of Hong Kong}

\affil[ ]{\tt\small liuzhuom@usc.edu, hding15@jhu.edu, wjli@ie.cuhk.edu.hk}
\vspace{-0.5cm}
\maketitle
\ificcvfinal\thispagestyle{empty}\fi

\begin{abstract}
   The existing active learning methods select the samples by evaluating the sample's uncertainty or its effect on the diversity of labeled datasets based on different task-specific or model-specific criteria. 
   In this paper, we propose the Influence Selection for Active Learning(ISAL) which selects the unlabeled samples that can provide the most positive Influence on model performance. 
   To obtain the Influence of the unlabeled sample in the active learning scenario, we design the Untrained Unlabeled sample Influence Calculation(UUIC) to estimate the unlabeled sample's expected gradient with which we calculate its Influence. 
   To prove the effectiveness of UUIC, we provide both theoretical and experimental analyses. 
   Since the UUIC just depends on the model gradients, which can be obtained easily from any neural network, our active learning algorithm is task-agnostic and model-agnostic. 
   ISAL achieves state-of-the-art performance in different active learning settings for different tasks with different datasets. 
   Compared with previous methods, our method decreases the annotation cost at least by 12\%, 13\% and 16\% on CIFAR10, VOC2012 and COCO, respectively.
\end{abstract}

\vspace{-0.5cm}
\section{Introduction}
\begin{figure}[!t]
   \centering
   \includegraphics[width=1\linewidth]{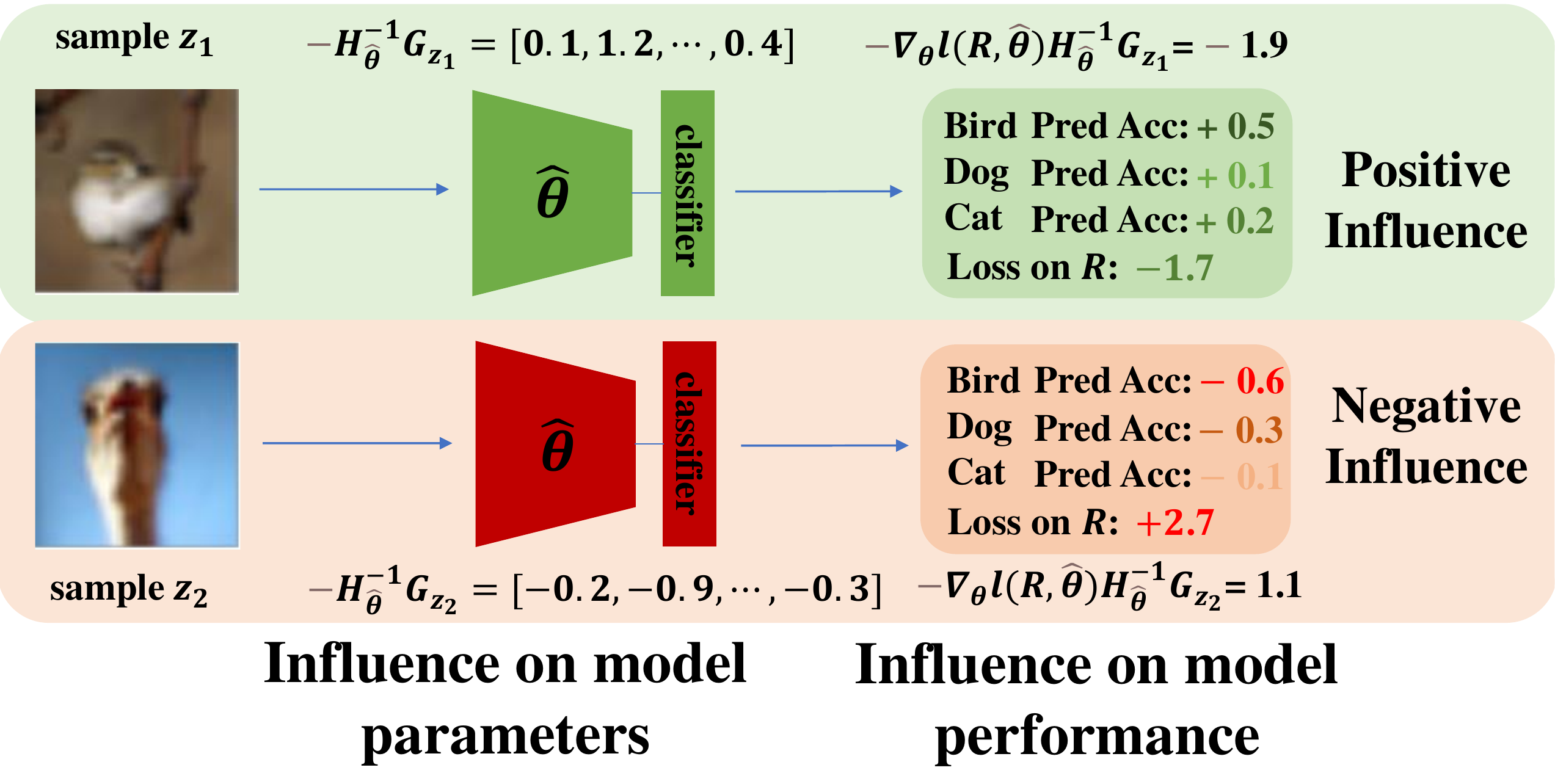} 
   \vskip -0.1in
   \caption{Using UUIC to calculate the influence of unlabeled samples. 
   These two samples will be annotated as 'Bird' if they are selected. 
   In UUIC, we calculate the influence of the sample by calculating $-\nabla_\theta l(R,\hat\theta)^T H_{\hat\theta}^{-1} G_{z_{i}}$.
   The more negative the influence value is, the more positive influence on model performance the sample provides.
   Base on the result from UUIC, our ISAL algorithm selects the sample $z_{1}$ for annotation.} 
   \label{influence} 
   \vskip -0.15in
 \end{figure}

 Active learning is a kind of sampling algorithm that aims to reduce the annotation cost by helping the model to achieve better performance with fewer labeled training samples. 
 In those areas with a limited annotation budget or the areas that need large amounts of labeled samples, active learning plays an important and irreplaceable role. 
 However, unlike the rapid progress of weakly supervised learning and semi-supervised learning, the development of active learning is limited.
 Especially in the computer vision area, most of the existing active learning algorithms are restricted to the image classification problem.
 
 Given a pool of unlabeled images, different active learning algorithms evaluate the importance of each image with different criteria, which
 can be divided into uncertainty-based methods and diversity-based methods. The uncertainty-based methods~\cite{lewis1994sequential,joshi2009multi,seung1992query,gal2016dropout,yoo2019learning} 
 use different criteria to evaluate the uncertainty of an image and select the images that the trained model is less confident about. 
 However, the neural network shows over-confidence~\cite{hein2019relu} toward the unfamiliar samples,
 indicating that using the uncertainty to estimate the samples' importance may not be accurate, deteriorating the performance of the active learning algorithm.
 
 The diversity-based methods~\cite{nguyen2004active,yang2015multi,hasan2015context,sener2017active} aim to select a subset from the whole unlabeled dataset with the largest diversity. 
 These methods do not consider the model state. Besides, some of them need to measure the distance between each labeled image and each unlabeled image, 
 meaning that their computation complexity are quadratic with respect to the size of the dataset. This disadvantage will become more apparent on the large-scale dataset.
 
 In addition to image classification, object detection is also an important area that has large amounts of applications. 
 The annotation for the datasets ~\cite{lin2014microsoft, shao2019objects365, pascal-voc-2012} of object detection is extremely time-consuming. 
 Thus, active learning for object detection is well demanded. 
 However, the research in active learning for object detection~\cite{roy2018deep,kao2018localization,desai2020towards,haussmann2020scalable} 
 is rare and most of the proposed methods are designed for specific architecture, e.g., Faster R-CNN~\cite{ren2015faster} or SSD~\cite{liu2016ssd}.
 
 

 In this paper, instead of designing a task or even architecture-specific algorithms, we propose an algorithm that can be generally applied to different tasks and architectures. 
 There are already some successful attempts like the diversity-based coreset~\cite{sener2017active} and the uncertainty-based learning loss~\cite{yoo2019learning} algorithm, 
 which proves that the general algorithm for active learning is possible. 
 Unlike these two algorithms that select samples by measuring the feature distance or the expected loss which are assumed to be correlated with the potential influence on the model, 
 our method estimates the samples' influence directly. 
 
 Our method, Influence Selection for Active Learning(ISAL), selects samples with the most positive influence, i.e. the model performance will be enhanced most by adding this sample with full annotation into the labeled dataset. 
 The influence measurement was first proposed by Cook~\cite{cook1982residuals} for robust statistics. 
 However, the scenario for the influence estimation in our work is entirely different. 
 In our case, the samples are unlabeled and untrained. 
 We design the Untrained Unlabeled sample Influence Calculation(UUIC) to calculate the influence of the unlabeled and untrained sample by estimating its expected gradient. 
 Figure.~\ref{influence} shows how UUIC evaluates unlabeled samples and helps ISAL select samples.
 Since UUIC just needs to use the model gradients, which can be easily obtained in a neural network no matter 
 what task is and how complex the model structure is, our proposed ISAL is task-agnostic and model-agnostic.
 
 ISAL achieves state-of-the-art performance among all comparing active learning algorithms for both the image classification and object detection task in the commonly used active learning setting with different representative datasets. 
 Our method saves 12\%, 13\%, 16\% annotation than the best comparing methods in CIFAR10~\cite{krizhevsky2009learning}, VOC2012~\cite{pascal-voc-2012} and COCO~\cite{lin2014microsoft}, respectively. 
 In addition, the existing methods for object detection perform better than random sampling only when the trained model's performance is far lower than the ones trained on the full dataset, indicating that some selected samples may not be the best choice. 
 Thus, we apply ISAL to a large-scale active learning setting for object detection. 
 ISAL decreases the annotation cost at least by 8\% than all comparing methods when the detector reaches 94.4\% performance of the model trained on the full COCO dataset.

 The contribution of this paper is summarized as follows:
 \begin{enumerate}
    \item
    We propose Influence Selection for Active Learning(ISAL), a task-agnostic and model-agnostic active learning algorithm, which selects samples based on the calculated influence. 
    \item
    We design the Untrained Unlabeled sample Influence Calculation(UUIC), a method to calculate the influence of the unlabeled and untrained sample by estimating its expected gradient. 
    To validate UUIC's effectiveness, we provide both theoretical and experimental analyses.
    \item
    ISAL achieves state-of-the-art performance in different experiment settings for both image classification and object detection.
 \end{enumerate}

\section{Related Work}
The existing active learning methods~\cite{ren2020survey} can be divided into two categories: 
uncertainty-based and diversity-based methods. 
Many of them are designed for image classification or can be used in classification without much change.

\textbf{Uncertainty-based Methods.} The uncertainty has been widely used in active learning to estimate samples’ importance. 
It can be defined as the posterior probability of a predicted class~\cite{lewis1994sequential, lewis1994heterogeneous, wang2016cost}, 
the posterior probability margin between the first and the second predicted class~\cite{joshi2009multi, roth2006margin}, 
or the entropy of the posterior probability distribution~\cite{settles2012active, joshi2009multi, luo2013latent, settles2008analysis}. 
In addition to directly using the posterior probability, researchers design some different methods for evaluating the samples’ uncertainty. 
Seung~\cite{seung1992query} trains multiple models to construct a committee and measures uncertainty by the consensus between the multiple predictions from the committee. 
Gal~\cite{gal2016dropout} proposes an active learning method that obtains uncertainty estimation through multiple forward passes with Monte Carlo Dropout. 
Yoo~\cite{yoo2019learning} creates a module that learns how to predict the unlabeled images' loss and chooses the unlabeled image with the highest predicted loss.
Freytag~\cite{freytag2014selecting} selects the images with the biggest expected model output changes, which can be also regarded as the uncertainty-based method.

\textbf{Diversity-based Method.} It aims to solve the sampling bias problem in batch querying. 
To achieve this goal, a clustering algorithm is applied~\cite{nguyen2004active} or a discrete optimization problem~\cite{yang2015multi, elhamifar2013convex, guo2010active} is solved.
The core-set approach~\cite{sener2017active} attempts to solve this problem by constructing a core subset. 
In addition to using k-Center-Greedy to calculate the core subset, its performance can be further enhanced by solving a mixed-integer program. 
The context-aware methods~\cite{hasan2015context, mac2014hierarchical} consider the distance between the samples and their surrounding points to enrich the diversity of the labeled dataset.
Sinha~\cite{sinha2019variational} trains a variational autoencoder and an adversarial network to discriminate between unlabeled and labeled samples, which can also be regarded as a diversity-based method.

\textbf{Active Learning for Object Detection.} The research in active learning for object detection is rare and most of the existing methods need complicated design. 
Roy~\cite{roy2018deep} selects the images with the biggest offset between the bounding boxes(bboxes) predicted in intermediate layers and the last layer of the SSD~\cite{liu2016ssd} model. 
Kao~\cite{kao2018localization} proposes to use the intersection over union(IoU) between the bboxes predicted by the Region Proposal Network(RPN) head and Region of Interest(RoI)
head of Faster R-CNN~\cite{ren2015faster} to measure the image uncertainty, or measuring the uncertainty of an image by the change of the predicted bboxes under different levels of data augmentation, and chooses the images with the highest uncertainty. 
Desai~\cite{desai2020towards} measures the bbox-level uncertainty and proposes a new method that chooses bboxes for active learning instead of images. 
Haussmann~\cite{haussmann2020scalable} examines different existing methods in the scenario of large-scale active learning for object detection.
He finds that the method which achieves the best performance chooses the images with more bboxes, increasing the annotation cost which is contradictory to the purpose of the active learning. 
In fact, most of the researches ignore that the annotation cost of object detection is closely relative to the bboxes number instead of the image number. 

\textbf{Influence Function.} Cook~\cite{cook1982residuals} first introduces influence function for robust statistics. 
The influence function evaluates the importance of a trained sample by measuring how the model parameters change as we upweight this sample by an infinitesimal amount. 
Recently, Koh~\cite{koh2017understanding} uses the influence function to understand the neural network model behavior. 
Ren~\cite{ren2020not} evaluates the trained unlabeled sample in semi-supervised learning by influence function. 
However, as far as we know, none of the existing publications uses the influence function on the untrained sample. 
Besides, Cook's derivation of influence function is also based on the trained sample. 
Thus there is no solid theoretical support for using the influence function on the untrained sample so far.

\section{Method}
In this section, we start with the problem definition of active learning. 
In Section~\ref{Evaluation of an untrained data}, we provide a derivation for evaluating the influence of an untrained sample. 
In Section~\ref{Evaluation of an untrained unlabeled data}, we introduce Untrained Unlabeled sample Influence Calculation(UUIC) to estimate an untrained unlabeled sample's expected gradient with which we calculate the influence of this sample. 
In Section~\ref{Influence Selection for Active Learning}, we show our proposed Influence Selection for Active Learning algorithm.


\subsection{Problem Definition} \label{Problem Definition}
In this section, we formally define the problem of active learning. 
We focus on some traditional computer vision tasks such as image classification and object detection. 

\begin{figure}[!t]
   \centering
   \includegraphics[width=1\linewidth]{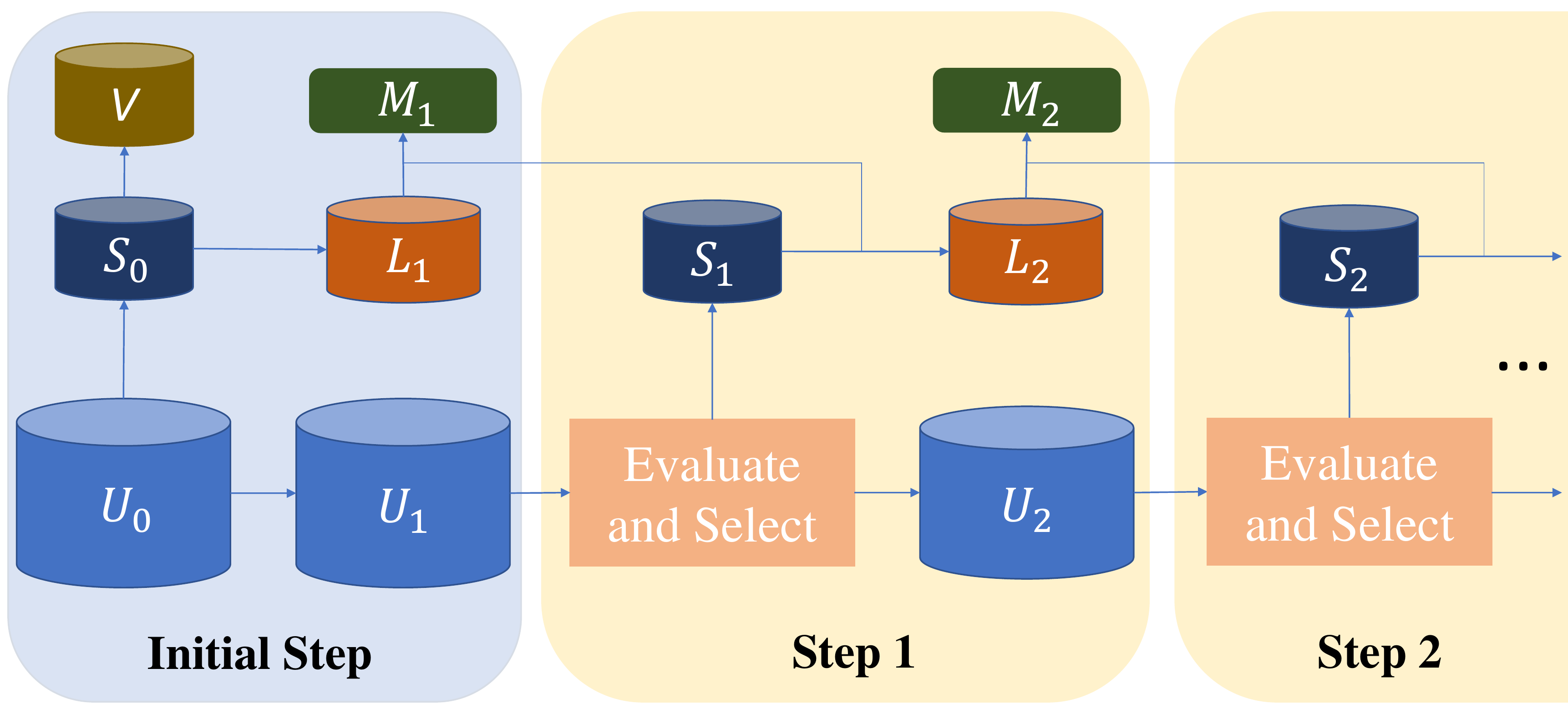} 
   \vskip -0.1in
   \caption{The pipeline of active learning. The iteration will be repeated until the model achieves a satisfactory performance or until we have exhausted the budget for annotation.} 
   \label{activeleanrning} 
   \vskip -0.1in
 \end{figure}

In real-world setting, we gather a large pool of unlabeled samples $U_{0}$. 
We randomly select a small amount of samples $S_{0}$ from $U_{0}$ and annotate them, the $U_{1}=U_{0} \setminus S_{0}$. 
The $S_{0}$ will be split into two parts, the initial labeled samples $L_{1}$, and the validation set $V$, which will be used to measure the trained model performance. 
The $L_{1}$ will be used to train the first model $M_{1}$ in active learning iteration. 
Then all unlabeled samples in $U_{1}$ will be evaluated according to some specific criteria. 
In our proposed method, we calculate the unlabeled sample's influence on model performance and use influence value as the criterion to evaluate the importance of the untrained sample. 
Based on the evaluation result, a new group of unlabeled samples $S_{1}$ will be selected and annotated. 
The labeled and unlabeled dataset will be updated, $U_{2} = U_{1} \setminus S_{1}$, and the $L_{2}=L_{1} \cup S_{1}$. 
Then $L_{2}$ will be used to train another model $M_{2}$, and $U_{2}$ will be evaluated and $S_{2}$ will be selected. 
This iteration will be repeated until the model achieves a satisfactory performance on $V$ or until we have exhausted the budget for annotation. 
Fig.~\ref{activeleanrning} shows the pipeline of active learning.

\subsection{The Influence of an Untrained Sample} \label{Evaluation of an untrained data}
In each step of the active learning, except the initial step, we have an unlabeled dataset $U_{i}$ and a labeled dataset $L_{i}$. 
With all the samples in $L_{i} = \{z_1, z_2, \cdots,z_n \}$ and loss function $\mathcal{L}(\theta)=\frac1n{\textstyle\sum_{j=1}^n}l(z_j,\theta)$, we train a model with its parameters $\theta\in\Theta$. 
The model would converge to $\hat\theta\in\Theta$, where $\hat\theta\overset{def}=\arg\min\;_{\theta\in\Theta}\;\frac1n{\textstyle\sum_{j=1}^n}l(z_j,\theta)$.

Next, we need to evaluate each unlabeled sample and select the most useful samples. We first measure the model parameters change due to adding a new sample $z^{'} \in U_{i}$ into the labeled dataset. We evaluate $z^{'}$ with the assumption that we already have its ground truth label. 
The change of the parameter is $\hat\theta_{z^{'}}-\hat\theta$, where $\hat\theta_{z^{'}}=\arg\min\;_{\theta\in\Theta}\;\frac1{n+1}\textstyle\sum_{z^{'}\cup L_i}l(z,\theta)$. 
However, retraining the model is time-consuming, and it's impossible to retrain a model for each unlabeled sample. Inspired by the motivation of the influence 
function~\cite{cook1982residuals}, we can compute approximation of the parameter change by adding a small influence from sample $z^{'}$ to the loss function $\mathcal{L}(\theta)$, giving us new parameters 
$\hat\theta_{\varepsilon,z^{'}}=\arg\min\;_{\theta\in\Theta}\;\frac1n{\textstyle\sum_{j=1}^n}l(z_j,\theta)+\varepsilon l(z^{'},\theta)$. 
Assuming that the loss function is twice-differentiable and strictly convex in $\theta$, the influence of sample $z^{'}$ on parameter $\hat\theta$ is given by 
\begin{equation}
   \label{eq:gradsimi}
   \begin{aligned}
   I(z^{'}) = {\left.\frac{\displaystyle\operatorname d\hat\theta_{\varepsilon,z^{'}}}{\displaystyle\operatorname d\varepsilon}\right|}_{\varepsilon=0} = -H_{\hat\theta}^{-1}\nabla_\theta l(z^{'},\hat\theta)
   \end{aligned}
\end{equation}
where $-H_{\hat\theta}^{-1} = \frac1n{\textstyle\sum_{j=1}^n\nabla_\theta^2}l(z_j,\hat\theta)$ is the Hessian and is positive definite by assumption.
See the supplementary material for the derivation in detail.

However, the model parameters change could not directly reflect the model performance change caused by the sample $z^{'}$. 
Thus, we randomly select and annotate a subset of unlabeled dataset $U_{0}$. This subset, named as reference set $R$, can 
represent the distribution of $U_{0}$. Next, we apply the chain rule to evaluate the influence of sample $z^{'}$ on model performance which 
is evaluated by the change of the model loss on reference set $R$:
\begin{equation}
   \label{eq:gradsimi2}
   \begin{aligned}
   I(z^{'},R) &= {\left.\frac{\displaystyle\operatorname dl(R,{\hat\theta}_{\varepsilon,z^{'}})}{\displaystyle\operatorname d\varepsilon}\right|}_{\varepsilon=0}\\
              &=\nabla_\theta l(R,\hat\theta)^T\;{\left.\frac{\displaystyle\operatorname d{\hat\theta}_{\varepsilon,z^{'}}}{\displaystyle\operatorname d\varepsilon}\right|}_{\varepsilon=0}\\
              &=-\nabla_\theta l(R,\hat\theta)^TH_{\hat\theta}^{-1} \nabla_\theta l(z^{'},\hat\theta)
   \end{aligned}
\end{equation}
The more negative $I(z^{'},R)$ is, the more positive on model performance influence $z^{'}$ can provide.
In practice, we select the validation set $V$ created in the first step of active learning as reference set, since this would not cause additional annotation. Our ablation study in Section~\ref{The possible substitute of reference set} shows, it's possible for us to use the labeled dataset as $R$ to calculate the $I(z^{'},R)$ in active learning, though using the validation set as reference set will perform better.
\begin{algorithm}
   \caption{Untrained Unlabeled sample Influence Calculation}\label{alg:gradexpectation}
   \begin{algorithmic}[1]
       \State \textbf{Input:} $s_{test}$, $z^{'}$
       \State Forward the $z^{'}$ into model $M_{i}$ 

       \State \textbf{if} Task is Image Classification \textbf{then}
       \State \quad Use the class with the highest posterior probability as $P$
       \State \textbf{else if} Task is Object Detection \textbf{then}
       \State \quad Filter the predicted bboxes with a given threshold   
       \State \quad Select the remaining bboxes as $P$ 
       \State \textbf{else} 
       \State \quad Generate the pseudo-label $P$ based on the task
       \State \textbf{end if}        

      \State Calculate the loss $\mathcal{L}_{z^{'}}=l(z^{'}, P, \hat\theta)$ 
      \State Back Propagate the $\mathcal{L}_{z^{'}}$ and get the $G_{z^{'}}$  
      \State \textbf{Return} $I(z^{'},R) = -s_{test} \cdot G_{z^{'}}$
   \end{algorithmic}
 \end{algorithm}

\subsection{Untrained Unlabeled sample Influence Calculation} \label{Evaluation of an untrained unlabeled data}
In our active learning setting, we need to evaluate an untrained sample $z^{'} \in U_{i}$ without the ground truth label. 
Therefore, we propose the Untrained Unlabeled sample Influence Calculation(UUIC) to calculate the influence of each sample in the unlabeled dataset.
Our aim is to measure the expected gradient $G_{z^{'}}$ of sample $z^{'}$ and to replace the $\nabla_\theta l(z^{'},\hat\theta)$ with $G_{z^{'}}$ in equation~\ref{eq:gradsimi2} for influence calculation.

We first focus on the expected gradient in image classification. 
The most intuitive design of the expected gradient is to use the top $K$ classes $\{label_0, label_1, \cdots, label_K\}$ in the posterior probability as ground truth label to calculating the loss. 
We backpropagate the losses to the model and obtain the gradients with respect to class $label_i$. 
Then, we use the posterior probability $pred_i$ of class $label_i$ as a weight to average the backpropagated gradients. 
The expected gradient $G_{z^{'}}$ is defined as 
\begin{equation}
   \label{eq:expected gradient}
   \begin{aligned}
      G_{z^{'}} = \sum_{i=1}^K\;\nabla_\theta l(z,label_i,\hat\theta)\cdot pred_i
   \end{aligned}
\end{equation}
Our experiments in~\ref{Expected gradient of untrained unlabeled data} shows that when $K$ is equal to 1, using the $G_{z^{'}}$ to calculate the unlabeled sample influence for active learning, our algorithm achieves the best performance. 
This indicates that we can use the pseudo-label $P$ as ground truth label to calculate the gradient of sample $z^{'}$ as $G_{z^{'}}$ in active learning.
We further apply this simple but effective way to calculate the $G_{z^{'}}$ in object detection, it also helps our active learning algorithm to achieve state-of-the-art performance.

After obtaining the $G_{z^{'}}$, we replace the $\nabla_\theta l(z^{'},\hat\theta)$ in equation~\ref{eq:gradsimi2} with it. Thus the influence of untrained unlabeled sample could be evaluated as
\begin{equation}
   \label{eq:gradsimi3}
   \begin{aligned}
   I(z^{'},R) = -\nabla_\theta l(R,\hat\theta)^T H_{\hat\theta}^{-1} G_{z^{'}}
   \end{aligned}
\end{equation}
Since in equation~\ref{eq:gradsimi3} the Hessian matrix $H_{\hat\theta}$ is symmetric, and $\nabla_\theta l(R,\hat\theta)$ and $G_{z^{'}}$ is a vector, the order of multiplication would 
not matter. In practice, we do not calculate the inverse matrix of the Hessian matrix. Instead we calculate the stochastic estimation~\cite{agarwal2016second} of Hessian-vector products $s_{test} = \nabla_\theta l(R,\hat\theta) H_{\hat\theta}^{-1}$ 
, which ensures that the computation complexity of our algorithm is $O(n)$. 
See the supplementary material for more implement details of $s_{test}$ calculation. 
After obtaining $s_{test}$, we calculate $I(z^{'},R) = -s_{test} \cdot G_{z^{'}}$. 
The algorithm~\ref{alg:gradexpectation} shows Untrained Unlabeled sample Influence Calculation. 

\subsection{Influence Selection for Active Learning} \label{Influence Selection for Active Learning}

The algorithm~\ref{alg:untainedunlabel} shows how Influence Selection for Active Learning algorithm uses UUIC to select samples from the unlabeled dataset.

\begin{algorithm}
   \caption{Influence Selection for Active Learning}\label{alg:untainedunlabel}
   \begin{algorithmic}[1]
       \State Compute the model gradient on reference set $\nabla_\theta l(R,\hat\theta)$
       \State Compute the $s_{test}$ with $\nabla_\theta l(R,\hat\theta)$
       \For{each sample $z^{'}$ in $U_{i}$}
         \State Compute the $I(z^{'},R)$ by algorithm~\ref{alg:gradexpectation} with input $s_{test}$, $z^{'}$
       \EndFor
       \State Sort all unlabeled samples base on $I(z^{'},R)$ 
       \State Select $|S_{i}|$ samples base on the active learning setting
   \end{algorithmic}
 \end{algorithm}

 \section{Experiment} 
 Since the active learning algorithm is a sampling algorithm, indicating that the performance of the algorithm may be sensitive to the dataset. 
 Therefore, we evaluate ISAL on different benchmarks in both classification and object detection to show its generalization ability and compare it with other methods that can be generally adapted to these tasks.  
 We further evaluate ISAL performance with the object detection dataset within a large-scale setting, which has not been mentioned before as far as we know. 
 Finally, we conduct the ablation study with visualization analysis. 
 
 The main experimental results have been provided as plots due to the limited space. 
 We provide tables in which we report the performance mean for each plot and implement details of all comparing methods in the supplementary material.
 
 \subsection{Image Classification} \label{Classification} 
 Image classification is the most common task which is used in the previous works to validate their methods. 
 In this task, the neural network model is trained to recognize the categories of the input images. 
 The category of the image needs to be labeled in the active learning task.
 
 \textbf{Datasets.}
 Both CIFAR10 and CIFAR100 contains 50000 images for training and 10000 images for testing. 
 SVHN has 73257 images for training, 26032 images for testing.
 We use the train set as an unlabeled set and evaluate the model performance on the test set. 
 We use classification accuracy as the evaluation metric.
 
 \textbf{Active Learning Settings.}
 For the experiments on CIFAR10, we randomly select 1000 images from the unlabeled set as the initial labeled dataset, and in each of the following steps, we add 1000 images to the labeled dataset. 
 For CIFAR100, we randomly select 5000 images from the unlabeled set first and add 1000 images in the following steps. 
 For SVHN, we randomly select 2\% of the unlabeled set as the initial labeled dataset, and we add the same number of images in each of the following steps. 
 We simulate 10 active learning steps and stop the active learning iteration. 
 We use the default data augmentation in pycls~\cite{Radosavovic2019}, which includes random flip and crop. 
 We normalize the images using the channel mean and standard deviation of the training set. We repeat the experiment 5 times.
 
 \textbf{Target Model.}
 We use ResNet-18~\cite{he2016deep} to verify our method, we implement the model and different active learning methods base on pycls. We train the model for 
 200 epochs with the mini-batch size of 128 and the initial learning rate of 0.1. After training 160 epochs, we decrease the learning rate to 0.01. The momentum and the weight decay are 0.9 and 0.0005 respectively. 
 
 \textbf{Implement Details.}
 For all datasets, we use all parameters in ResNet-18 to calculate the influence, and we use the test set as reference set. 
 When calculating the $s_{test}$, we random sample 250 images from the labeled set.
 We repeatedly calculate the $s_{test}$ 4 times and use the value after averaging. We compare our method with random sampling, coreset sampling~\cite{sener2017active} and learning loss sampling~\cite{yoo2019learning}. 
 
\begin{figure*}[htbp]
   \vspace{-0.8cm}
   \centering
   \subfigure[]{
   \includegraphics[width=6.1cm]{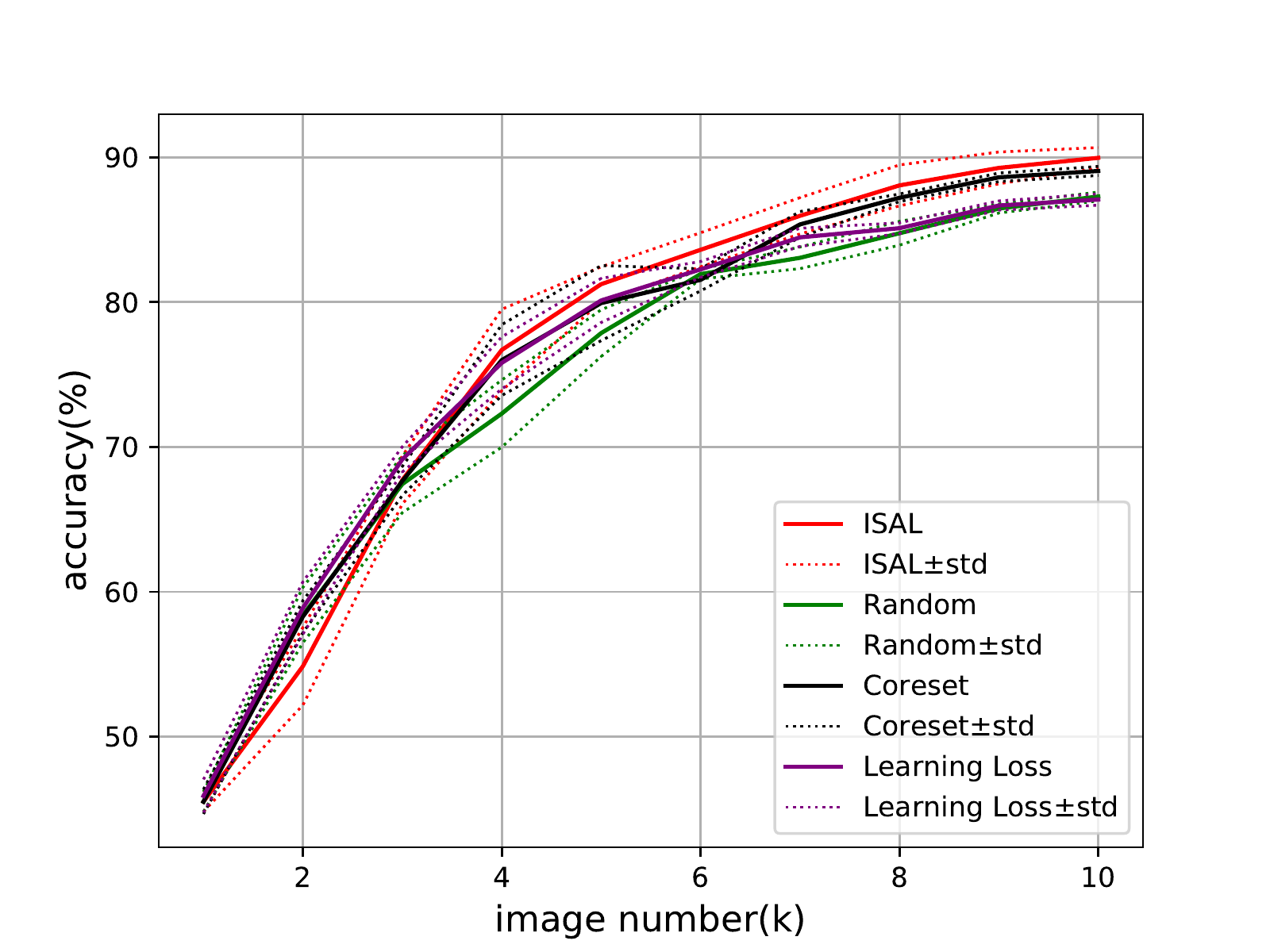}
   \label{CIFAR10}
   }
   \hspace{-9.5mm}
   \subfigure[]{
   \includegraphics[width=6.1cm]{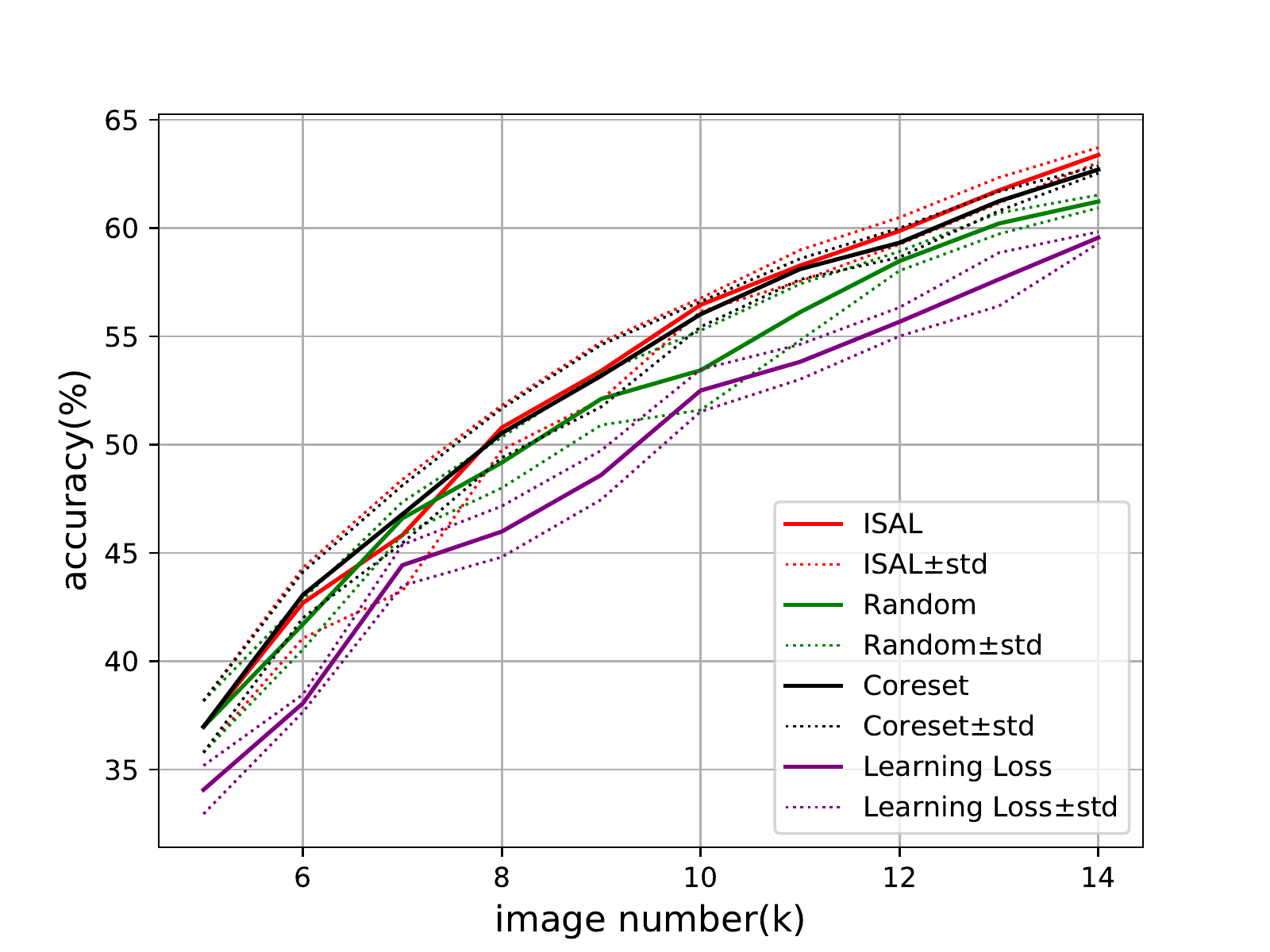}
   \label{cifar100}
   }
   \hspace{-9.5mm}
   \subfigure[]{
   \includegraphics[width=6.1cm]{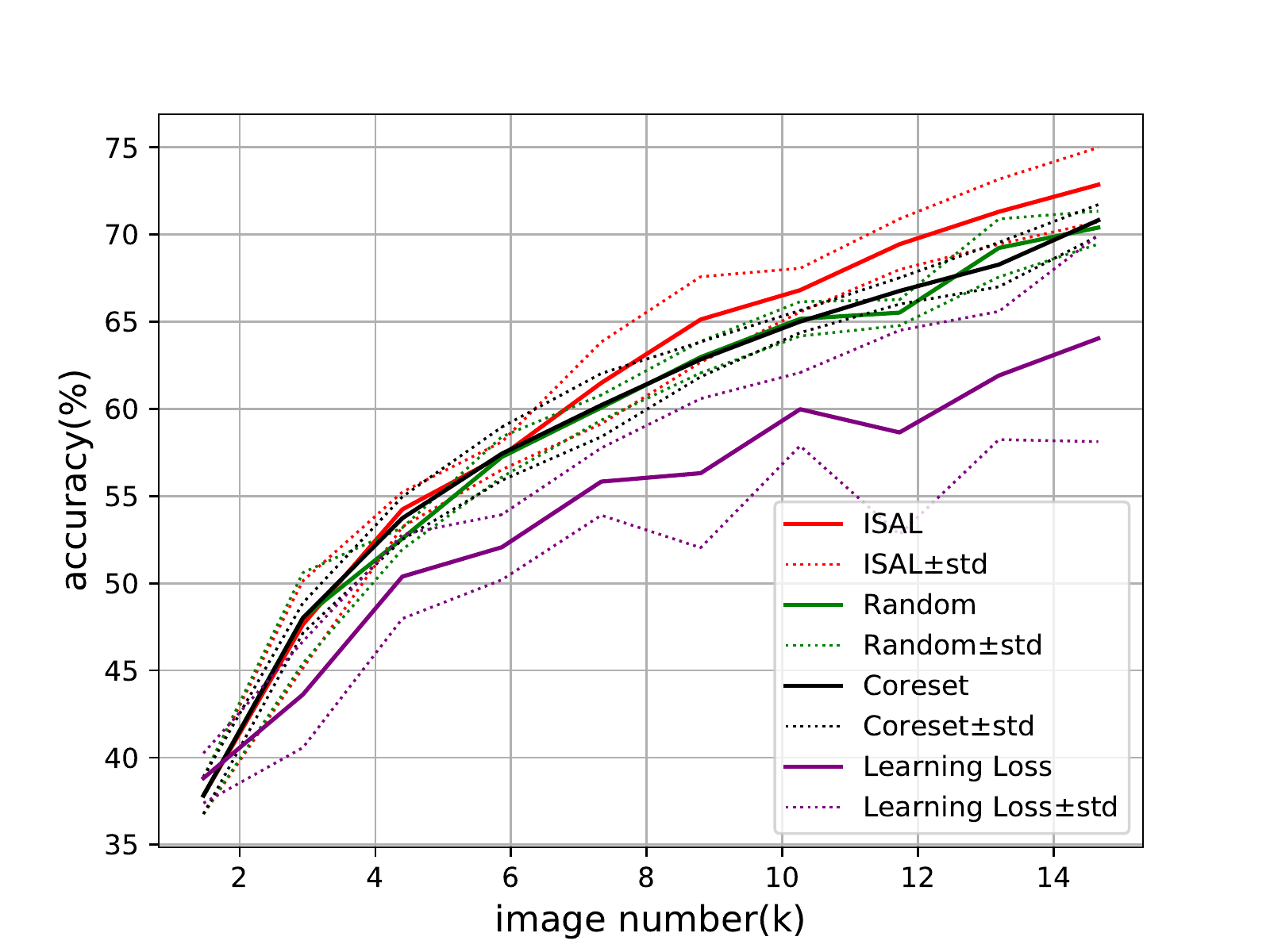}
   \label{SVHN}
   }
   \vspace{-4mm}
   \caption{Result for Image Classification. (a) Result on CIFAR10. (b) Result on CIFAR100. (c) Result on SVHN.}
   \vspace{-0.5cm}
\end{figure*}

 \textbf{Results.}
The results on CIFAR10, CIFAR100 and SVHN are shown in Figure~\ref{CIFAR10}, Figure~\ref{cifar100} and Figure~\ref{SVHN} respectively. 
We show how much annotations our method can save when it reaches other methods' final performance, the trained model performance after 10 active learning iterations.
For CIFAR10, our method uses roughly 1200 images fewer than the coreset sampling when achieving the final performance of coreset sampling, saving 12\% of annotation. 
When comparing with random sampling, our method saves roughly 2300 images when achieving the final performance of random sampling, saving 23\% of annotation.
For CIFAR100, our method uses roughly 400 and 1300 images fewer than the coreset sampling and random sampling, saving 2.9\% and 9.3\% of annotation respectively.
For SVHN, our method uses roughly 1800 and 2100 images fewer than the coreset sampling and random sampling, saving 12\% and 14\% of annotation respectively.

 \subsection{Object Detection} \label{Object Detection} 
 Object detection aims to detect instances of semantic objects of a certain class in images. The detectors are trained to localize the object by drawing bounding boxes(bboxes) and classifying the object inside the bounding box. The bboxes need to be drawn for the specific classes and the category of the object in the bboxes need to be annotated in the active learning task. 
 In practice, we found that the annotation cost of each image differs largely from others. Take COCO dataset as an example, the image in it at most has 63 bboxes and at least has zero bboxes. Thus, the cost of annotating a set of images is highly correlated with the number of bboxes instead of the number of images. 
 Thus, in the following experiments for object detection, we plot the average number of the bounding box and average mAP/AP from three tries.

 \textbf{Datasets.} 
 We choose the VOC2012~\cite{pascal-voc-2012}, which has been widely used in other active learning methods for object detection~\cite{kao2018localization, yoo2019learning}, 
 and COCO~\cite{lin2014microsoft}, a dataset that is commonly used to evaluate the performance of a detector.
 VOC2012 has 5717 images for training and 5823 images for validation, we use the trainset as the unlabeled dataset and use the validation set to evaluate the trained model performance. We use the mAP as the evaluation metric. 
 COCO dataset has 118k images for training and 5000 images for validation. We use the trainset as the unlabeled dataset, and validate the model performance on the validation set. We use AP as the evaluation metric. We use the default data preparation pipeline which includes the random flipping, 
 image normalization with channel mean and standard deviation, image resizing, and padding from the mmdetection~\cite{chen2019mmdetection}.

 \textbf{Active Learning Settings.}
 For the experiments on VOC2012, we randomly select 500 images from the unlabeled set as the initial labeled dataset, and in each of the following steps of the active learning cycle, we add the 500 images to the labeled set. We simulate 10 active learning iteration steps. 
 For COCO, we randomly select 5000 images from the unlabeled set first and add 1000 images in the following step. 
 Since the number of bounding boxes selected by different methods has huge differences, for clearer comparison, 
 we continue the active learning iteration until the trained model achieves $22\pm0.3\%$ in AP.

 \textbf{Target Model.}
 We use FCOS~\cite{tian2019fcos} detector with backbone ResNet-50 implemented in mmdetection to verify our method. 
 We also implement the active learning pipeline and different active learning methods base on mmdetection. We train the model for 12 epochs with the mini-batch size of 8 and the initial learning rate of 0.01.
 After training 8 and 11 epochs, we decrease the learning rate by 0.1 respectively. The momentum and the weight decay are 0.9 and 0.0001 respectively.
 
 \textbf{Implement Details.}
 For both datasets, when calculating the influence of the unlabeled data, we backpropagate the loss to the parameters in FCOS's last convolution layer, which contains three kernels used to generate the final prediction of classification, regression, and centerness score. 
 We use the validation set as reference set. 
 When calculating the $s_{test}$, we random sample at most 5000 images from the labeled set. We repeatedly calculate the $s_{test}$ 4 times and use the value after averaging. 
 We compare our method with random sampling, coreset sampling~\cite{sener2017active}, learning loss sampling~\cite{yoo2019learning} and localization stability sampling~\cite{kao2018localization}. 
 
 \begin{figure*}[htbp]
   \vspace{-0.8cm}
   \centering
   \subfigure[]{
   \includegraphics[width=6.1cm]{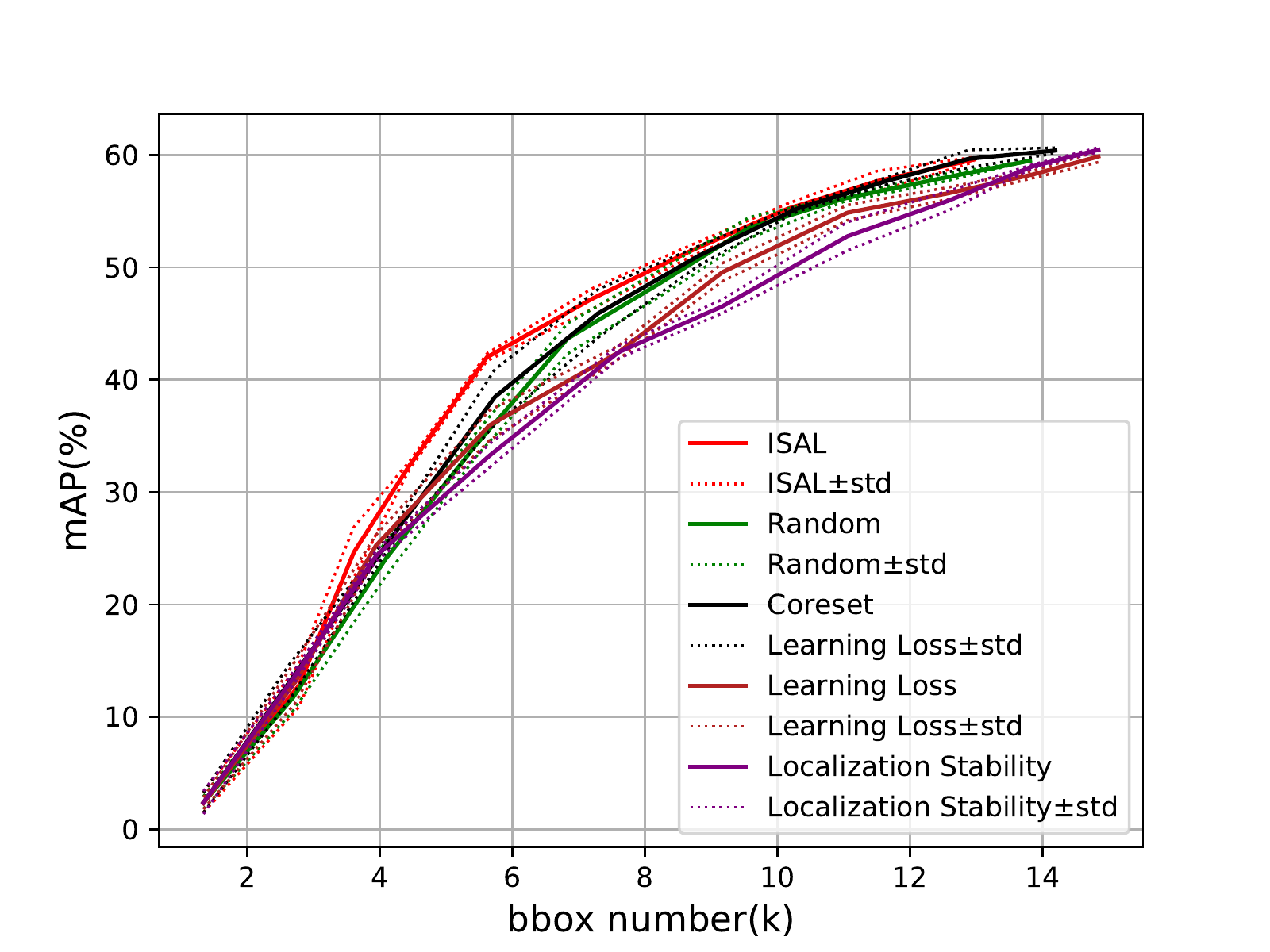}
   \label{voc2012}
   }
   \hspace{-9.5mm}
   \subfigure[]{
   \includegraphics[width=6.1cm]{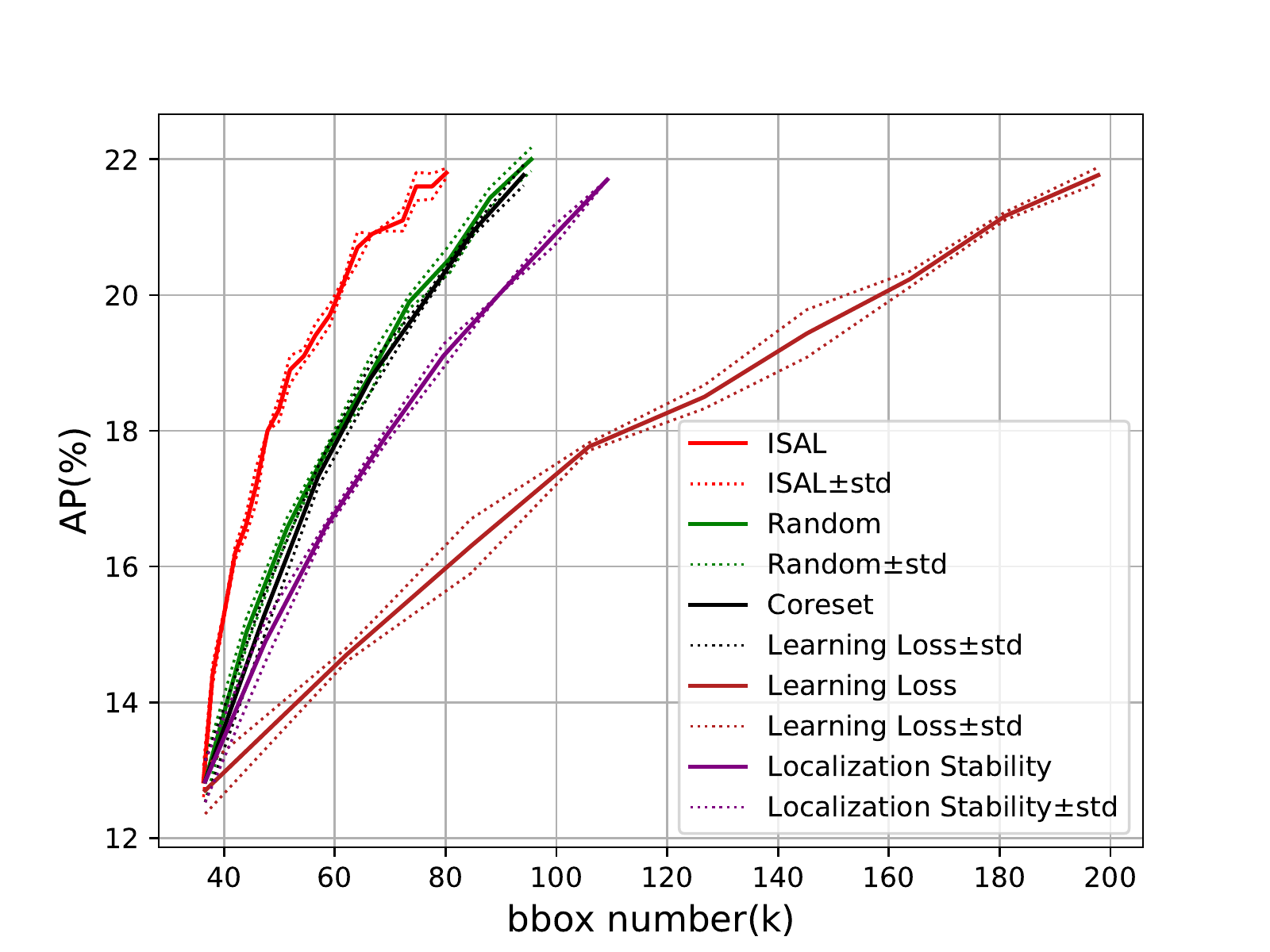}
   \label{coco}
   }
   \hspace{-9.5mm}
   \subfigure[]{
   \includegraphics[width=6.1cm]{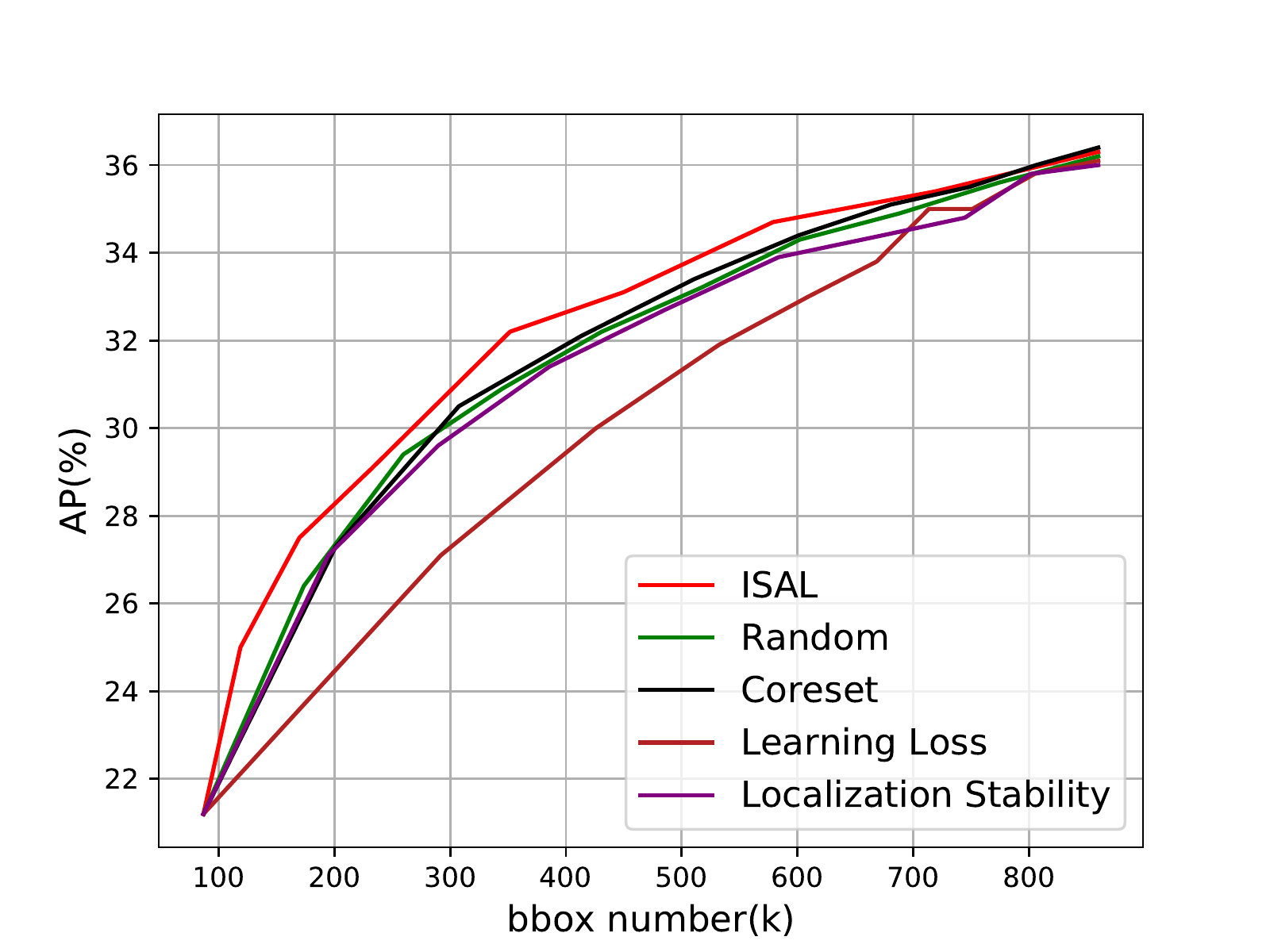}
   \label{cocofull}
   }
   \vspace{-4mm}
   \caption{Result for Object Detection. (a) Result on VOC2012. (b) Result on COCO. (c) Result on COCO within a large-scale setting.}
   \vspace{-0.5cm}
\end{figure*}

 \textbf{Results.}
 The result on VOC2012 and COCO are shown in Figure~\ref{voc2012} and Figure~\ref{coco} respectively. 
 For VOC2012 dataset, when the trained model achieves 42\% in mAP, our method uses roughly 850 bboxes fewer than coreset sampling, 
 saving 13\% of annotations, and saves roughly 2000 bboxes than localization stability sampling, decreasing the annotations by 26\%.
 Since the VOC2012 only has less than 6000 images, in the last iteration of active learning, different methods have selected similar images. 
 Thus, all methods achieve similar performance. Our method becomes more effective when it is applied to a large dataset.
 
 For COCO, when achieving the target AP, it costs 15.3k fewer bounding boxes than random and 117k fewer bounding boxes than the learning loss sampling, saving 16\% and 59\% annotation respectively.
 Our implementations show that all comparing methods perform worse than random sampling, meaning that their reported performance enhancement over random sampling is mainly caused by selecting the image with more bounding boxes. 
 Choose these images significantly enhance the annotation cost, which is contradictory to the purpose of active learning.

 \subsection{Large Scale Experiment in Object Detection}
 In this section, we conduct experiments on the large-scale active learning setting for object detection.
 It aims to prove that our method can be effective when the trained model performance is close to the performance of the model trained on the full dataset. 
 This experiment further validates the superiority of ISAL which can precisely select the samples with the most positive influence on model performance.

 \textbf{Datasets and Experiment Details.} 
 We use the COCO trainset as the unlabeled dataset, and validate the trained model performance on the validation set. 
 We plot the number of the bounding boxes and AP to show the model performance. 
 We randomly select 10\% images from the unlabeled set first and add the same number of images as the first step in the following steps. We iterate an active learning pipeline 10 times.
 We continue to use the FCOS detector with backbone ResNet-50 implemented in mmdetection to verify our method. 
 All other experiment details are the same as we described in Section~\ref{Object Detection}. 
 
 \textbf{Results.}
 The result is shown in Figure~\ref{cocofull}. When the trained model performance achieve 34\% in AP, which is close to the performance of the 
 FCOS trained on full COCO dataset, our method uses roughly 40k bounding boxes fewer than the coreset sampling, 
 which has the best performance in all comparing methods, decreasing the annotation cost by 8\%. 
 This also indicates that achieving 94.4\% performance of the model trained on full COCO dataset, we just need 60.7\% of the annotations of the dataset.
 
 \subsection{Ablation Study}
 In this section, to validate the effectiveness of UUIC and ISAL, we conduct experiments to discuss the properties of each element in $-\nabla_\theta l(R,\hat\theta)^T H_{\hat\theta}^{-1} G_{z^{'}}$.
 We conduct all the ablation studies on CIFAR10. All the experiment details are the same as mentioned in section~\ref{Classification}.
 
 \vspace{-3mm}
 \subsubsection{The Effect of K in Expected gradient} \label{Expected gradient of untrained unlabeled data}
 \vspace{-3mm}
 In this section, we discuss the effect of $K$ in the expected gradient $G_{z^{'}}$ with which we calculate the influence of the unlabeled sample. 
 Tab.~\ref{hyperparameter} shows the results. 
 When $K$ is equal to 1, our active learning algorithm achieves the best result in each step. 
 Our analysis shows that, in some cases, the direction of the gradient vector computed with the label of the first predicted class is just the opposite of the one computed with the label of the second predicted class. 
 Therefore, when averaging the gradient, some value in the $G_{z^{'}}$ will be diminished, making the influence of $z^{'}$ inaccurate.
 
 \begin{table}[!t]
    \centering
    \begin{tabular}{|c|ccccc|}
      \hline 
      \multirow{2}{*}{$K$} & \multicolumn{5}{c|}{number of CIFAR10 images} \\ \cline{2-6} 
                           & 1000 & 3000 & 5000 & 7000 & 9000 \\ \hline 
       \hline
      1&\textbf{45.52}&\textbf{67.72}&\textbf{81.24}&\textbf{85.96}&\textbf{89.26} \\
      2&45.52&65.65&81.20&85.37&88.15 \\
      5&45.52&65.19&78.52&85.43&89.13 \\
      10&45.52&62.26&72.28&80.14&83.03 \\
      \hline     
    \end{tabular}
    \centering
    \vspace{-3mm}
    \caption{The effect of $K$ in $G_{z^{'}}$.}
    \label{hyperparameter}   
 \end{table}
 
 \vspace{-3mm}
 \subsubsection{The Effect of $H_{\hat\theta}^{-1}$} \label{Components of the influence}
 \vspace{-3mm}
 In this section, we discuss the effect of $H_{\hat\theta}^{-1}$. 
 We compare the performance of ISAL with Gradient Similarity. 
 They use $-\nabla_\theta l(R,\hat\theta)^T H_{\hat\theta}^{-1} G_{z^{'}}$ and $-\nabla_\theta l(R,\hat\theta)^T G_{z^{'}}$ to evaluate and select the unlabeled samples, respectively.
 $-\nabla_\theta l(R,\hat\theta)^T G_{z^{'}}$ measures the similarity of gradients on reference set and the expected gradients of an untrained and unlabeled samples.
 
 Tab.~\ref{table:Hessianmatrix} shows that Gradient Similarity has a similar performance as ISAL, though the ISAL performs better. 
 In essence, the Gradient Similarity uses the gradients on the reference set to evaluate which parameters in the model have not been learned well and 
 selects the unlabeled images with a similar expected gradient to train in the next step. 
 This will help the model to obtain the biggest backpropagated gradients on specific model parameters, moving to the global optimal quickly. 
 However, some unlabeled images with different expected gradients also provide a positive influence on the model. A similar phenomenon is mentioned in ~\cite{koh2017understanding}.
 $H_{\hat\theta}^{-1}$ helps ISAL to find these samples and enhances ISAL performance.

 \begin{table}[!t]
    \centering
    \begin{tabular}{|c|ccccc|}
      \hline 
      \multirow{2}{*}{Method} & \multicolumn{5}{c|}{number of CIFAR10 images} \\ \cline{2-6} 
                           & 1000 & 3000 & 5000 & 7000 & 9000 \\ \hline 
       \hline
       ISAL&\textbf{45.52}&\textbf{67.72}&\textbf{81.24}&\textbf{85.96}&\textbf{89.26}\\
       Grad Simi&45.52&67.54&80.54&85.72&88.60\\ 
       \hline 
      \hline     
    \end{tabular}
    \centering
    \vspace{-3mm}    
    \caption{The effect of $H_{\hat\theta}^{-1}$ on the performance of ISAL.}
    \label{table:Hessianmatrix}   
 \end{table}

 \vspace{-3mm}
 \subsubsection{The Selection of Reference Set} \label{The possible substitute of reference set}
 \vspace{-3mm}
 In this section, we try different substitutes for using the validation set as the reference set. We try using the $L_{1}$ as the reference set in each step of the iteration, named as ISAL\_v2, and using the labeled dataset of each step $L_{i}$ as the reference set, named as ISAL\_v3.
 
 Tab.~\ref{table:referenceset} shows that the ISAL\_v2 and ISAL\_v3 performance is slightly worse than the ISAL, but they still perform much better than random sampling. 
 In essence, the gradients on the reference set represent whether the model parameters have fit in with the data distribution or not.
 Thus, to ensure that the calculated influence value can precisely represent the model performance change, the distribution of the reference set needs to be similar to the distribution of the $U_{0}$.
 Since the $L_{1}$ is also randomly sampled from $U_{0}$, the performance of ISAL\_v2 is more close to ISAL than ISAL\_v3. 
 However, $L_{1}$ has been trained. 
 The model gradients on $L_{1}$ become smaller than the gradients on the validation set, and the calculated influence value may not be precise, explaining why ISAL\_v2 performs worse than ISAL.

 \begin{table}[!t]
    \centering
    \begin{tabular}{|c|ccccc|}
      \hline 
      \multirow{2}{*}{Method} & \multicolumn{5}{c|}{number of CIFAR10 images} \\ \cline{2-6} 
                           & 1000 & 3000 & 5000 & 7000 & 9000 \\ \hline 
       \hline
       ISAL    &\textbf{45.52}&\textbf{67.72}&\textbf{81.24}&\textbf{85.96}&\textbf{89.26} \\
       ISAL\_v2&45.52&67.06&80.57&85.71&88.92 \\
       ISAL\_v3&45.52&67.12&80.11&84.88&88.71 \\ \hline \hline 
       coreset      &45.52&67.66&79.93&85.36&88.61\\
       random       &45.52&67.55&77.77&83.09&86.50\\
      \hline     
    \end{tabular}
    \centering
    \vspace{-3mm}    
    \caption{Comparision of different reference set.}
    \label{table:referenceset}   
 \end{table}

\subsection{Visualization Analysis}
\begin{figure}[!t]
   \centering
   \vspace{-4mm}
   \includegraphics[width=0.9\linewidth]{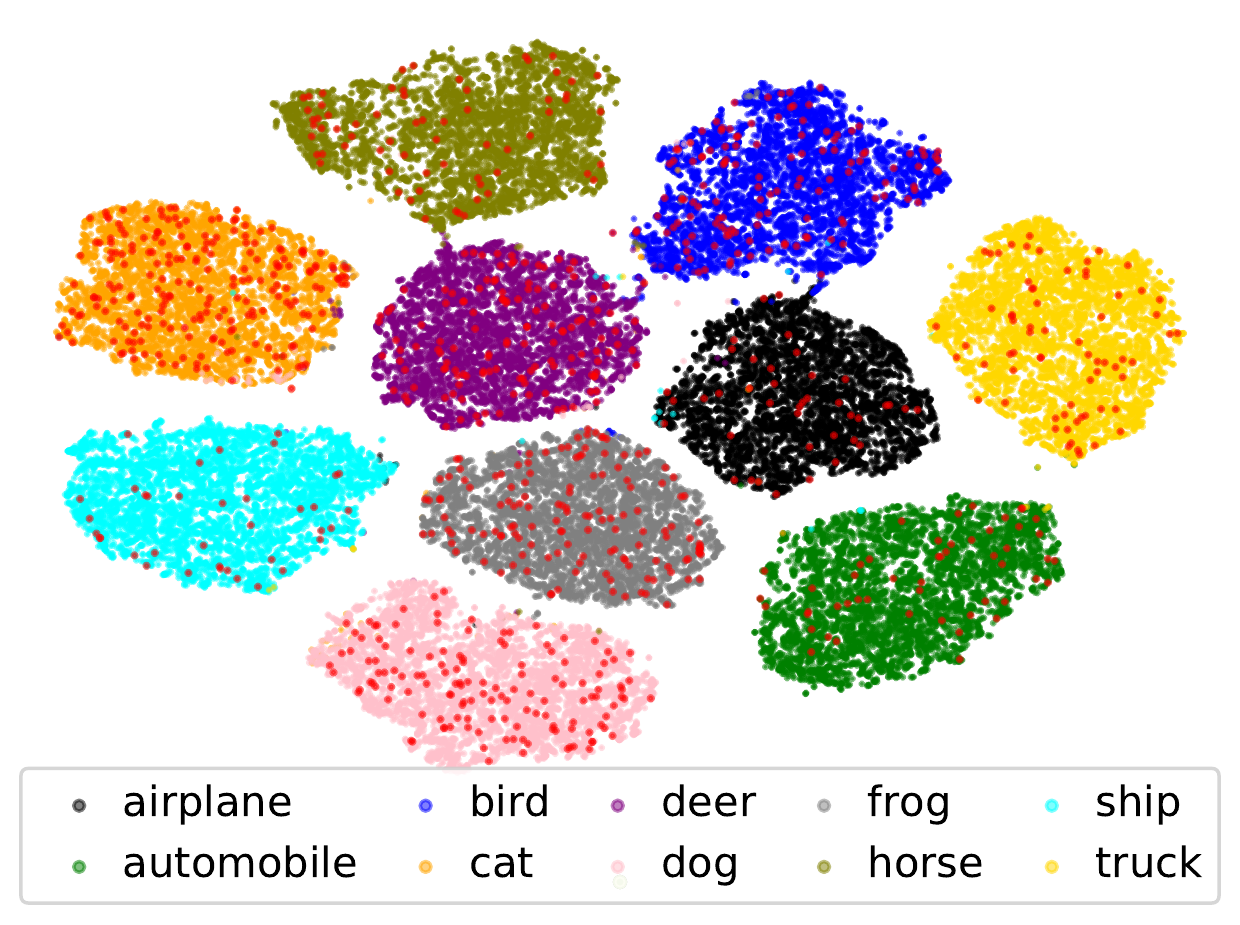} 
   \vskip -0.1in
   \caption{The tSNE embeddings of the CIFAR10 training set. 
   The red dots represent the images in $S_{1}$ selected by ISAL.} 
   \label{influence_dist} 
   \vskip -0.1in
 \end{figure}

Figure.~\ref{influence_dist} shows the tSNE embeddings of the CIFAR10 training set. The red dots represent the images in $S_{1}$ selected by ISAL.
Our proposed method tends to choose more images with cat, bird, and deer. 
Our analysis shows that $M_{1}$ has lower accuracy in these three classes.
Thus selecting the images of these three classes can provide a more positive influence on the model performance. 
In addition, the $M_{1}$ is trained on $L_{1}$ which is randomly sampled, but the model performs worse in these three classes than the other, indicating that these three classes are hard to learn. 
Thus, evenly sampling images from all classes would lead to data redundancy.
Instead, our proposed method selects samples in bias enhancing the learning efficiency.

\begin{figure}[!t]
   \centering
   \includegraphics[width=1\linewidth]{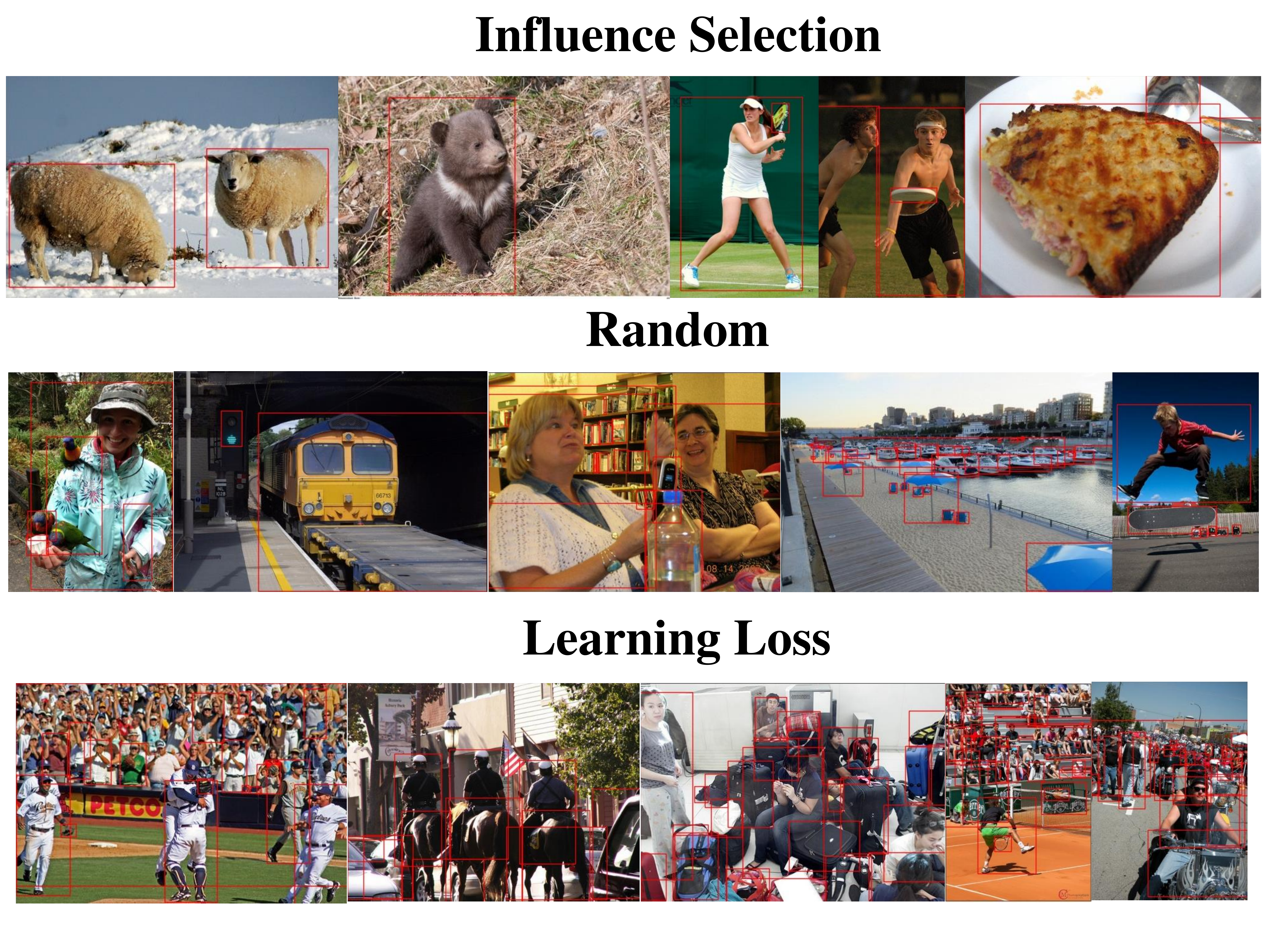} 
   \vskip -0.1in
   \caption{The selected images in COCO dataset by different active learning algorithms.} 
   \label{selected_image} 
   \vskip -0.1in
 \end{figure}

 Figure.~\ref{selected_image} shows some selected images of COCO dataset in $S_{1}$ by different methods.
 Our proposed method selects images with fewer bboxes, while the bboxes' size in the selected images is significantly larger than the one selected by other methods.
 In addition, the bboxes in the selected images of our proposed method have a lower overlap ratio. 
 This indicates that the clear and large object in the image helps the model learn more effectively.
 In the latter of the iteration, our proposed method will select the images with more objects and more complex scenarios, this would help the model to learn from the easy to the difficult step by step.
 
\section{Conclusion}
We have proposed a task-agnostic and model-agnostic active learning algorithm, Influence Selection for Active Learning(ISAL), helping neural networks model to learn more effectively and decreasing the annotation cost.
By making use of the Untrained Unlabeled sample Influence Calculation(UUIC) to calculate the influence value for each unlabeled sample, ISAL selects the samples which can provide the most positive influence on model performance.
ISAL achieves state-of-the-art performance on different tasks in both commonly use settings and a newly-designed large-scale setting.
We believe that ISAL can be extended to solve many active learning problems in other areas, and it would not be restricted to the tasks in computer vision.

\noindent\textbf{Acknowledgement:} We thank Zheng Zhu for implementing the classification pipeline, Bin Wang and Xizhou Zhu for helping with the experiments, and thank Yuan Tian and Jiamin He for discussing the mathematic derivation.

\clearpage

{\small
\bibliographystyle{ieee_fullname}
\bibliography{egbib}
}
\begin{center}
\section*{Supplementary Material}
\end{center}
\appendix
In this supplementary material, we provide additional details which we could not include in the main paper due to space constraints. The material is composed as follows:
\begin{enumerate}
   \item
   The derivation of the influence of untrained samples. 
   \item
   The implementation details of $s_{test}$ calculation.
   \item
   The time complexity analysis.
   \item
   The implementation details of comparing methods.
   \item
   Additional experiments on CIFAR10 dataset and COCO dataset.
   \item
   The tables in which we report the average performance for each plot.
\end{enumerate}

\section{The Derivation of the Influence of Untrained Samples} \label{derivationofinfluence}
\textbf{Newton Step and Quadratic Approximation.}
Assuming that we have labeled dataset $L_{i}$ and loss function $\mathcal{L}(\theta)=\frac1n\textstyle\sum_{z\in L_i}l(z,\theta)$.
After training a model on $L_{i}$, we have the model parameters $\hat\theta\in\Theta$, where $\hat\theta=\arg\min\;_{\theta\in\Theta}\;\frac1n\sum_{z\in L_i}l(z,\theta)$.
Our purpose is to estimate the parameters of model which is trained on $L_{i}$ and the new added sample $z^{'}$. 
The new loss function is $\mathcal{L}_{z^{'}}(\theta)=\frac1{n+1}\textstyle\sum_{z^{'}\cup L_i}l(z,\theta)$, 
giving new trained model parameter $\hat\theta_{z^{'}}=\arg\min\;_{\theta\in\Theta}\;\frac1{n+1}\textstyle\sum_{z^{'}\cup L_i}l(z,\theta)$. 

Considering the quadratic approximation of the $\mathcal{L}_{z^{'}}(\hat\theta_{z^{'}})$ 
\begin{equation}
   \label{eq:taylor_expension}
   \begin{aligned}
      \mathcal{L}_{z^{'}}(\hat\theta_{z^{'}}) = \mathcal{L}_{z^{'}}(\hat\theta) + (\hat\theta_{z^{'}}-\hat\theta)^{T} \nabla_\theta \mathcal{L}_{z^{'}}(\hat\theta) + \\
      \frac12(\hat\theta_{z^{'}}-\hat\theta)^{T} \nabla_\theta^{2} \mathcal{L}_{z^{'}}(\hat\theta) (\hat\theta_{z^{'}}-\hat\theta)
   \end{aligned}
\end{equation}
If the $H_{\hat\theta}^{-1}$ is positive definite, the quadratic approximation is minimized at
\begin{equation}
   \label{eq:newtonstep}
   \begin{aligned}
      \hat\theta_{z^{'}}-\hat\theta &= -\frac{\nabla_\theta \mathcal{L}_{z^{'}}(\hat\theta)}{\nabla_\theta^{2} \mathcal{L}_{z^{'}}(\hat\theta)} \\
      &= -[\nabla_\theta^{2} \mathcal{L}_{z^{'}}(\hat\theta)]^{-1}[\nabla_\theta \mathcal{L}_{z^{'}}(\hat\theta)]
   \end{aligned}
\end{equation}
Thus, the quadratic approximation of $\hat\theta_{z^{'}}$ is equal to $\hat\theta-[\nabla_\theta^{2} \mathcal{L}_{z^{'}}(\hat\theta)]^{-1}[\nabla_\theta \mathcal{L}_{z^{'}}(\hat\theta)]$, 
and $-[\nabla_\theta^{2} \mathcal{L}_{z^{'}}(\hat\theta)]^{-1}[\nabla_\theta \mathcal{L}_{z^{'}}(\hat\theta)]$ is the newton step.

\textbf{Evaluate the Influence of an Untrained Sample.}
First we add a small influence from $z^{'}$ to the loss function $\mathcal{L}(\theta)$, the new loss function is 
\begin{equation}
   \label{eq:deviation1}
   \begin{aligned}
      \mathcal{L}_{\varepsilon,z^{'}}(\theta) &= \arg\min\;_{\theta\in\Theta}\;\frac1n\sum_{z\in L_i}l(z,\theta) + \varepsilon l(z^{'},\theta) \\
                                   &= \mathcal{L}(\theta) + \varepsilon l(z^{'},\theta)
   \end{aligned}
\end{equation}
With new loss function, the model new parameters is obtained $\hat\theta_{\varepsilon,z^{'}}=\arg\min\;_{\theta\in\Theta}\;\mathcal{L}_{\varepsilon,z^{'}}(\theta)$.
We evaluate a sample $z^{'}$ importance by calculating the ${\left.\frac{\displaystyle\operatorname d\hat\theta_{\varepsilon,z^{'}}}{\displaystyle\operatorname d\varepsilon}\right|}_{\varepsilon=0}$

From equation~\ref{eq:newtonstep} we know that 
\begin{equation}
   \label{eq:deviation2}
   \begin{aligned}
      \hat\theta_{\varepsilon,z^{'}} - \hat\theta =& -[\nabla_\theta^{2}\mathcal{L}_{\varepsilon,z^{'}}(\hat\theta)]^{-1} [\nabla_\theta\mathcal{L}_{\varepsilon,z^{'}}(\hat\theta)]\\
                                   =& -[\nabla_\theta^{2}\mathcal{L}(\hat\theta) + \varepsilon \nabla_\theta^{2} l(z^{'},\hat\theta)]^{-1} \\
                                   & \qquad [\nabla_\theta\mathcal{L}(\hat\theta) + \varepsilon \nabla_\theta l(z^{'},\hat\theta)]
   \end{aligned}
\end{equation}
Since $\hat\theta$ minimizes $\mathcal{L}(\theta)$, $\nabla_\theta\mathcal{L}(\hat\theta)$ is equal to 0. Dropping the $O(\varepsilon^{2})$ terms, we have
\begin{equation}
   \label{eq:newlossfunction2}
   \begin{aligned}
      \hat\theta_{\varepsilon,z^{'}} - \hat\theta \approx -[\nabla_\theta^{2}\mathcal{L}(\hat\theta)]^{-1} \varepsilon \nabla_\theta l(z^{'},\hat\theta)
   \end{aligned}
\end{equation}
We define $H_{\hat\theta}^{-1} \overset{def}=[\nabla_\theta^{2}\mathcal{L}(\hat\theta)]^{-1}$, and we have

\begin{equation}
   \label{eq:newlossfunction4}
   \begin{aligned}
      \hat\theta_{\varepsilon,z^{'}} - \hat\theta \approx -H_{\hat\theta}^{-1} \varepsilon \nabla_\theta l(z^{'},\hat\theta)
   \end{aligned}
\end{equation}

Thus, we can evaluate a untrained sample by:
\begin{equation}
   \label{eq:newlossfunction3}
   \begin{aligned}
      {\left.\frac{\displaystyle\operatorname d\hat\theta_{\varepsilon,z^{'}}}{\displaystyle\operatorname d\varepsilon}\right|}_{\varepsilon=0} &= {\left.\frac{\hat\theta_{\varepsilon,z^{'}} - \hat\theta}{\displaystyle\operatorname \varepsilon}\right|}_{\varepsilon=0} \\
      &= {\left.\frac{-\varepsilon H_{\hat\theta}^{-1} \nabla_\theta l(z^{'},\hat\theta)}{\displaystyle\operatorname \varepsilon}\right|}_{\varepsilon=0} \\
      &= -H_{\hat\theta}^{-1} \nabla_\theta l(z^{'},\hat\theta)
   \end{aligned}
\end{equation}

\section{The Implementation Details of $s_{test}$ Calculation} \label{the implement details of calculation of influence function}
To evaluate an untrained unlabeled sample, $I(z^{'},R) = -\nabla_\theta l(R,\hat\theta)^T H_{\hat\theta}^{-1} G_{z^{'}}$ needs to be calculated. 
However, it's impossible to calculate the inverse matrix of the Hessian matrix due to the memory constrain of GPU and the time complexity, especially for the deep neural network. 
We use the method proposed by Agarwal~\cite{agarwal2016second} to effectively approximate the $s_{test} \overset{def}= H_{\hat\theta}^{-1} \nabla_\theta l(R,\hat\theta)$
and then calculate $I(z^{'},R) = -s_{test} \cdot G_{z^{'}}$ for each samples. 

Dropping the $\hat\theta$ subscript for clarity, we define 
\begin{equation}
   \label{eq:inversehession}
   \begin{aligned}
      H_{j}^{-1}\overset{def}=\textstyle\sum_{i=0}^j(I-H)^i
   \end{aligned}
\end{equation}
as the first j terms in the Taylor expansion of $H^{-1}$. When $j\rightarrow\infty$, we have $H_{j}^{-1} \rightarrow H^{-1}$. 

From equation~\ref{eq:inversehession}, we have 
\begin{equation}
   \label{eq:inversehession2}
   \begin{aligned}
      H_{j}^{-1} = I + (I - H)H_{j-1}^{-1}
   \end{aligned}
\end{equation}
The key idea of stochastic estimation is that we can substitute the full $H$ in equation~\ref{eq:inversehession2} with the any unbiased estimator of $H$ to form $\tilde H_j$. Since $\mathbb{E}[\tilde H_j^{-1}]=H_{j}^{-1}$, we still have
$\mathbb{E}[\tilde H_{j}^{-1}]=H^{-1}$, when $j\rightarrow\infty$. In practice, we can randomly sample $z_{i}$ and use $\nabla_\theta^{2}l(z_{i},\hat\theta)$ as
the unbiased estimator of $H$. Algorithm~\ref{alg:stest} shows how we approximate the $s_{test}$.

\begin{algorithm}
   \caption{The calculation of $s_{test}$}\label{alg:stest}
   \begin{algorithmic}[1]
       \State \textbf{Input:} $v = \nabla_\theta l(R,\hat\theta)$
       \State Random sample $k$ images $\{z_1, z_2,\cdots,z_k\}$ from labeled dataset
       \State initial the $s_{test_0} = v$
       \For{$i$ in range($1$, $k+1$)}
         \State $s_{test_i}= v + (I - \nabla_\theta^{2} l(z_i,\hat\theta)) s_{test_{i-1}}$
       \EndFor  
      \State take the $s_{test_k}$ as the unbiased estimator of $s_{test}$
      \State \textbf{Return} $s_{test}$
   \end{algorithmic}
 \end{algorithm}

 In practice, we calculate the Hessian-vector products of $\nabla_\theta^{2} l(z_i,\hat\theta) s_{test_{i-1}}$ instead of calculating the Hessian matrix $\nabla_\theta^{2} l(z_i,\hat\theta)$. 
 We will repeat the algorithm~\ref{alg:stest} $p$ times, and use the averaged result as the final estimation of $s_{test}$.

 

 
\section{The Time Complexity Analysis}
As demonstrated in Section~\ref{the implement details of calculation of influence function}, our method can be divided into two sections. 
First, instead of directly calculate the $H_{\hat\theta}^{-1}$, we sample images from the labeled dataset to calculate the $s_{test}$, 
which is the stochastic estimation of $\nabla_\theta l(R,\hat\theta)^T H_{\hat\theta}^{-1}$.
Since the number of sampled images is fixed, the time complexity is a constant $C$.
Then, we calculate the influence for each unlabeled sample with $s_{test}$. 
Noted that $|U|=n$, the time complexity is $O(n)$.

\section{The Implementation Details of Comparing Methods} \label{The Implement Details of Comparing Methods}
\subsection{Image Classification}
 For coreset sampling~\cite{sener2017active}, we follow~\cite{yoo2019learning} and implement the K-Ceter-Greedy algorithm, which is just slightly worse than the mixed-integer program but much less time-consuming. 
 We run the algorithm by using the feature before the classification layer as~\cite{sener2017active} reported. 
 For the learning loss sampling, we connect the learning loss module to each block of ResNet-18, stopping the loss prediction module gradient from back-propagating to the model after 120 epochs, 
 and set the $\lambda$ to 1 as~\cite{yoo2019learning} do. 
 We first randomly select a subset with 10000 images from unlabeled samples before predicting the loss and selecting the image with the largest predicted loss.

 \subsection{Object Detection} \label{sup_od}
 For coreset sampling, we implement the K-Ceter-Greedy algorithm. 
 We apply global average pooling on the feature after the regression branch and the classification branch of FCOS~\cite{tian2019fcos}, then we concatenate the features from both branches and use this to run the algorithm. 
 We also tried using the feature from the Feature Pyramid Network(FPN) of FCOS to run the algorithm, but it does not perform better.
 
 For the learning loss sampling, we use the 5 feature maps from the FPN of FCOS. 
 We stopping the loss of the loss prediction module from back-propagating to the backbone, otherwise, the detector performance would deteriorate significantly. 
 We set the $\lambda$ to 1. 

 For localization stability sampling~\cite{kao2018localization}, we implement the Localization Stability method in the paper, since its performance is evaluated on both VOC2012~\cite{pascal-voc-2012} and COCO~\cite{lin2014microsoft} datasets.
 \subsection{Large Scale Experiment in Object Detection}
 All the implementation details of the comparing methods are exactly the same as~\ref{sup_od}

 \begin{figure}[!t]
   \centering
   \includegraphics[width=1\linewidth]{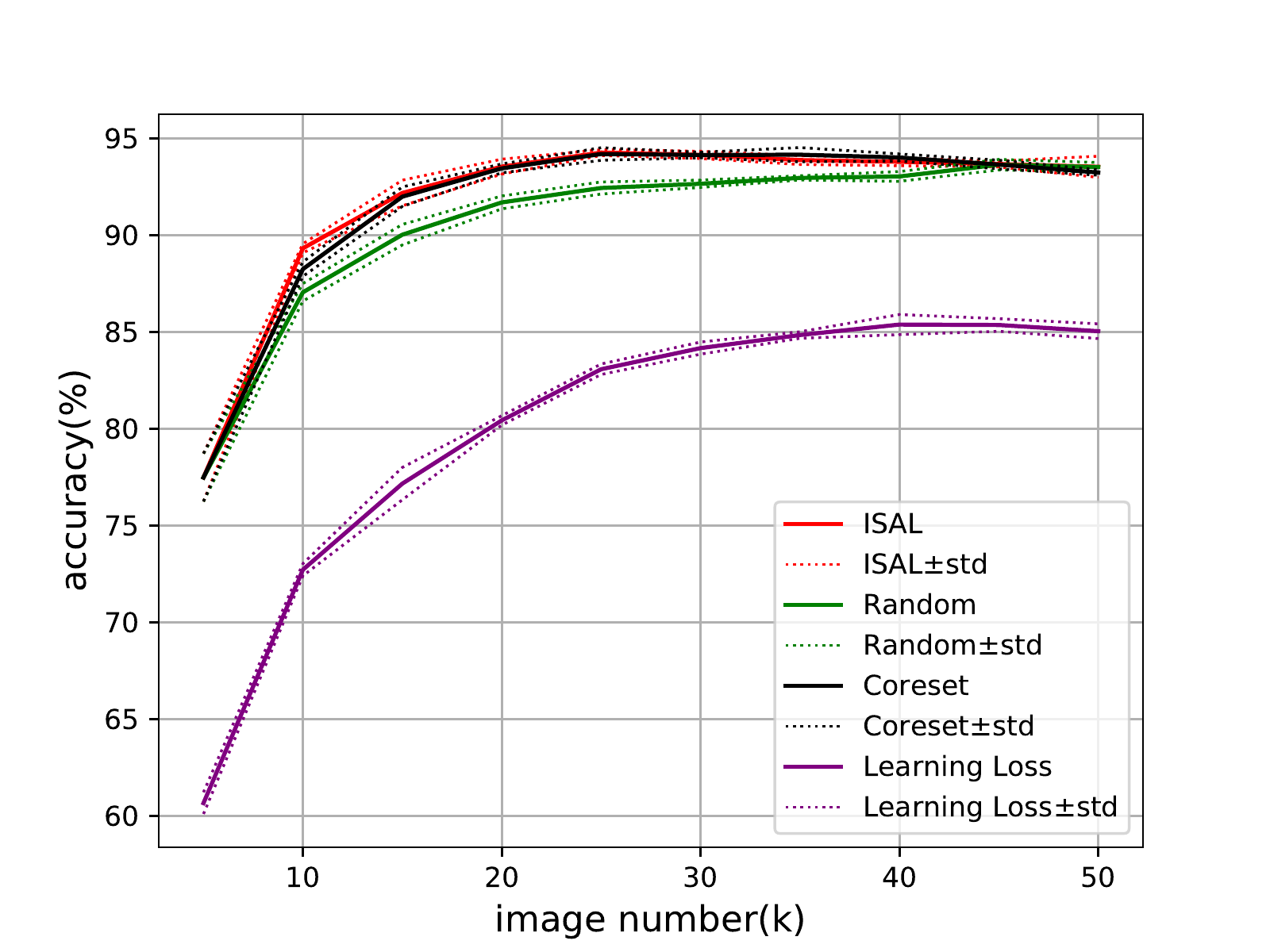} 
   \vskip -0.1in
   \caption{Result for CIFAR10 in large-scale active learning setting.} 
   \label{cifar10largescale} 
   \vskip -0.1in
 \end{figure}

 \begin{figure}[!t]
   \centering
   \includegraphics[width=1\linewidth]{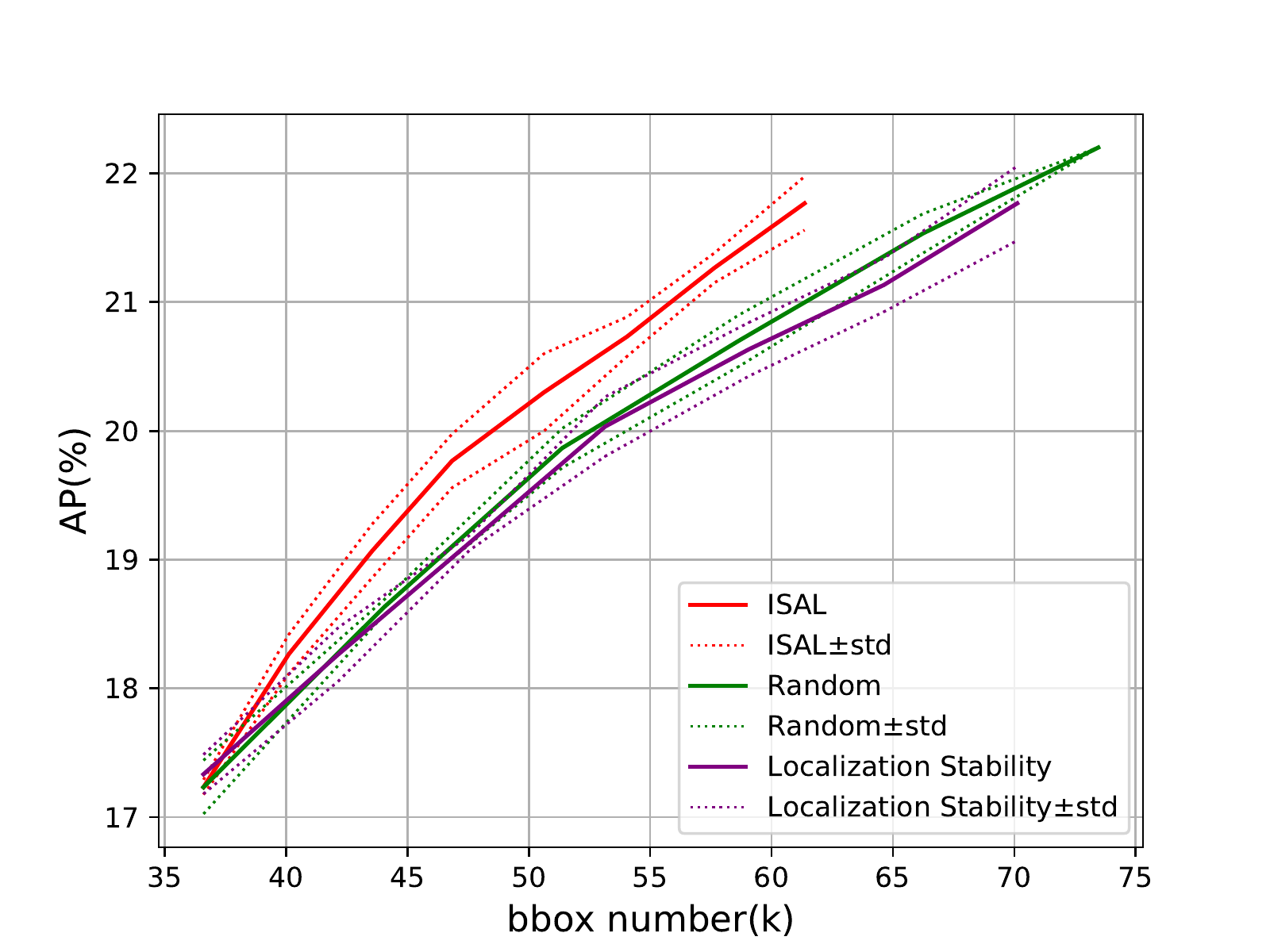} 
   \vskip -0.1in
   \caption{Result for COCO with Faster R-CNN.} 
   \label{fasterrcnn} 
   \vskip -0.1in
 \end{figure}

 \section{Additional Experiments}
 \subsection{Image Classification} \label{additional_class}
 In this section, we provide additional experiments on image classification with CIFAR10 in large scale active learning setting.

 \textbf{Active Learning Settings.}
 For the experiments on CIFAR10, we randomly select 5000 images from the unlabeled set as the initial labeled dataset, and in each of the following steps, we add 5000 images to the labeled dataset. 
 The simulate 10 active learning steps and stop the active learning iteration. 
 All other implementation details are exactly the same as we described in the main paper.

 \textbf{Results.} The results on CIFAR10 with large-scale active learning setting are shown in Figure~\ref{cifar10largescale}. 
 Our proposed method outperforms all comparing methods before step 6. 
 Our implementation shows that both our method and coreset sampling achieve the best performance at step 5, and the performance of the trained model deteriorates when we keep enlarging the labeled dataset.
 In practice, it is not necessary to continue the active learning iteration after step 5.
 This phenomenon indicates that, when using the ResNet-18 as the classifier and using the test set of CIFAR10 as the benchmark to evaluate the model performance, some images in the training set of CIFAR10 provide a negative influence on the model's performance.
 Active learning algorithm does help trained model to achieve better performance with fewer annotations.

 \subsection{Object Detection} \label{additional_det}
 In this section, we provide additional experiments on object detection with the COCO dataset.

 \textbf{Active Learning Settings.}
 We randomly select 5000 images from the unlabeled set first and add 1000 images in the following steps. 
 Since the number of bounding boxes selected by different methods has huge differences, for clearer comparison, 
 we continue the active learning iteration until the trained model achieves $22\pm0.3\%$ in AP.

 \textbf{Target Model.}
 We use Faster R-CNN~\cite{ren2015faster} detector with backbone ResNet-50 implemented in mmdetection~\cite{chen2019mmdetection} to verify our method. 
 We train the model for 12 epochs with the mini-batch size of 8 and the initial learning rate of 0.01.
 After training 8 and 11 epochs, we decrease the learning rate by 0.1 respectively. The momentum and the weight decay are 0.9 and 0.0001 respectively.
 
 \textbf{Implementation Details.}
 When calculating the influence of the unlabeled data, we backpropagate the loss to the parameters in the last convolution layer for regression and classification in Region Proposal Network(RPN),
 and to fully connected layer for regression result and classification result in Region of Interest Network(RoI) of Faster R-CNN. 
 We use the validation set as reference set. 
 When calculating the $s_{test}$, we random sample at most 500 images from the labeled set. We repeatedly calculate the $s_{test}$ 4 times and use the value after averaging. 
 We compare our method with random sampling and localization stability sampling~\cite{kao2018localization}, which can be implemented in Faster R-CNN easily. 
 For localization stability sampling~\cite{kao2018localization}, we implement the Localization Stability method in the paper.

 \textbf{Results.}
The results on Faster R-CNN are shown in Figure~\ref{fasterrcnn}. 
When achieving $21.8$ in AP, our method cost 7.1k fewer bounding boxes than random sampling, saving 10.4\% annotations.
This result shows that our method can be effective in both one-stage and two-stage detectors. 
It further substantiates that our method is task-agnostic and model-agnostic.

 \section{The Experiment Results}
 Table~\ref{experiment:CIFAR10}, table~\ref{experiment:CIFAR100} and table~\ref{experiment:SVHN} show the experiment results on the image classification of the main paper.
 Table~\ref{experiment:CIFAR10largescale} shows the experiment result of Section~\ref{additional_class} in supplementary material.

 Table~\ref{experiment:voc2012}, table~\ref{experiment:coco} and table~\ref{experiment:cocofull} show the experiment results on the object detection of the main paper.
 Table~\ref{experiment:coco_fasterrcnn} shows the experiment result of Section~\ref{additional_det} in supplementary material.

 \begin{table*}[!t]
   \centering
   \begin{tabular}{|c|ccccc|}
     \hline 
     \multirow{2}{*}{Methods} & \multicolumn{5}{c|}{5 times average of Accuracy(\%) in each step} \\ \cline{2-6} 
                          & 1 & 2 & 3 & 4 & 5  \\ \hline 
      \hline
     ISAL         &45.51799931           &54.86599902           & 67.72399840          & \textbf{76.69599825} & \textbf{81.23799817} \\
     coreset      &45.51799931           &58.29599852           & 67.65599847          & 75.99799808          &79.93399816 \\
     random       &45.51799931           &58.40199852           & 67.44199847          & 72.30399829          &77.86199819 \\
     learningloss & \textbf{45.91599973} & \textbf{58.88999852} & \textbf{69.14799823} & 75.80599807          &80.11599825 \\
     
     \hline \hline    
     \multirow{2}{*}{Methods} & \multicolumn{5}{c|}{5 times average of Accuracy(\%) in each step} \\ \cline{2-6} 
                          & 6 & 7 & 8 & 9 & 10 \\ \hline 
      \hline
     ISAL         & \textbf{83.61199812} & \textbf{85.95799826} &\textbf{88.05599827} &\textbf{89.26399810} &\textbf{89.95799797} \\
     coreset      & 81.53599801          & 85.36399841           &87.19999817          &88.61399807         &89.05199809 \\
     random       & 81.93599806          & 83.05799810           &84.75199825          &86.45999833         &87.28999833 \\
     learningloss & 82.25999810          & 84.46999836           &85.10199790          &86.66799833         &87.07999842 \\
     
     \hline     

   \end{tabular}
   \centering

   \vspace{-3mm}
   \caption{The experiment results on CIFAR10 with ResNet-18.}
   \vspace{-3mm}  
   \label{experiment:CIFAR10}   
\end{table*}

\begin{table*}[!t]
   \centering
   \begin{tabular}{|c|ccccc|}
     \hline 
     \multirow{2}{*}{Methods} & \multicolumn{5}{c|}{5 times average of Accuracy(\%) in each step} \\ \cline{2-6} 
                          & 1 & 2 & 3 & 4 & 5  \\ \hline 
      \hline
     ISAL         &\textbf{36.97799921} &42.69599870           &45.82199922          &\textbf{50.78799950} & \textbf{53.40799696} \\
     coreset      &\textbf{36.97799921} &\textbf{43.06599873}  &\textbf{46.78799956} &50.53399960         &53.17999910 \\
     random       &\textbf{36.97799921} &41.70599863           &46.59999991          &49.17400009         &52.11799930 \\
     learningloss &34.06799937          &38.06399904           &44.43599916          &45.98999956         &48.60199980 \\
     
     \hline \hline
     \multirow{2}{*}{Methods} & \multicolumn{5}{c|}{5 times average of Accuracy(\%) in each step} \\ \cline{2-6} 
                          & 6 & 7 & 8 & 9 & 10 \\ \hline 
      \hline
     ISAL         & \textbf{56.45799872} &\textbf{58.26799854} &\textbf{59.87199865} &\textbf{61.7459986} &\textbf{63.37799866} \\
     coreset      & 56.02399869          &58.10799861          &59.32599477          &61.24399836         &62.70399857 \\
     random       & 53.43599904          &56.11799880          &58.47799854          &60.21399841         &61.22999834 \\
     learningloss & 52.50199925          &53.82599889          &55.67399864          &57.63399866         &59.55999863 \\
     
     \hline    
   \end{tabular}
   \centering

   \vspace{-3mm}
   \caption{The experiment results on CIFAR100 with ResNet-18.}
   \vspace{-3mm}  
   \label{experiment:CIFAR100}   
\end{table*}

\begin{table*}[!t]
   \centering
   \begin{tabular}{|c|ccccc|}
     \hline 
     \multirow{2}{*}{Methods} & \multicolumn{5}{c|}{5 times average of Accuracy(\%) in each step} \\ \cline{2-6} 
                          & 1 & 2 & 3 & 4 & 5  \\ \hline 
      \hline
     ISAL         &37.82575205          &47.63829102           &\textbf{54.23709167} &57.31100053         &\textbf{61.49200840} \\
     coreset      &37.82575205          &\textbf{48.01090959}  &53.73232837          &\textbf{57.43315776}&60.21665482 \\
     random       &37.82575205          &48.01090931           &52.56760825          &57.24646455         &60.07759528 \\
     learningloss &\textbf{38.82068125} &43.61785414           &50.38337374          &52.06561057         &55.83128330 \\
     
     \hline \hline
     \multirow{2}{*}{Methods} & \multicolumn{5}{c|}{5 times average of Accuracy(\%) in each step} \\ \cline{2-6} 
                          & 6 & 7 & 8 & 9 & 10 \\ \hline 
      \hline
     ISAL         & \textbf{65.13214355} &\textbf{66.81315140} &\textbf{69.44836955} &\textbf{71.31146101}&\textbf{72.86339711} \\
     coreset      & 62.84495852          &65.01152269          &66.76705426          &68.27980783         &70.82436816 \\
     random       & 62.97172561          &65.16671630          &65.52473728          &69.23401795         &70.40181142 \\
     learningloss & 56.31453489          &59.98309629          &58.65165838          &61.91687003         &64.04348344 \\
     
     \hline    
   \end{tabular}
   \centering

   \vspace{-3mm}
   \caption{The experiment results on SVHN with ResNet-18.}
   \vspace{-3mm}  
   \label{experiment:SVHN}   
\end{table*}

\begin{table*}[!t]
   \centering
   \begin{tabular}{|c|ccccc|}
     \hline 
     \multirow{2}{*}{Methods} & \multicolumn{5}{c|}{5 times average of Accuracy(\%) in each step} \\ \cline{2-6} 
                          & 1 & 2 & 3 & 4 & 5  \\ \hline 
      \hline
     ISAL         &\textbf{77.48999786}&\textbf{89.33399824}&\textbf{92.19199778}&\textbf{93.55399844}&\textbf{94.27999855} \\
     coreset      &\textbf{77.48999786}&88.25799831         &92.00199783&93.46399852&94.20599862 \\
     random       &\textbf{77.48999786}&87.05999821         &90.03599803&91.70399761&92.44999793 \\
     learningloss &60.66199856         &72.72399856         &77.17199846&80.43999806&83.08399811 \\
     
     \hline \hline
     \multirow{2}{*}{Methods} & \multicolumn{5}{c|}{5 times average of Accuracy(\%) in each step} \\ \cline{2-6} 
                          & 6 & 7 & 8 & 9 & 10 \\ \hline 
      \hline
     ISAL         &\textbf{94.15999845}&93.89199833         &93.80199861         &\textbf{93.68999855}&\textbf{93.54599838} \\
     coreset      &94.15399850         &\textbf{94.16999856}&\textbf{94.02599862}&93.66199836         &93.24799830 \\
     random       &92.66799801         &92.97199826         &93.04599803         &93.65399836         &93.53199820 \\
     learningloss &84.17599796         &84.85199794         &85.39199788         &85.36999783         &85.04999776 \\
     
     \hline       
   \end{tabular}
   \centering

   \vspace{-3mm}
   \caption{The experiment results on CIFAR10 in large-scale setting with ResNet-18.}
   \vspace{-3mm}  
   \label{experiment:CIFAR10largescale}   
\end{table*}

 \begin{table*}[!t]
   \centering
   \begin{tabular}{|c|c|ccccc|}
     \hline 
     \multicolumn{2}{|c|}{\multirow{2}*{Method}} & \multicolumn{5}{c|}{3 times average of results in each step} \\ \cline{3-7} 
     \multicolumn{2}{|c|}{~}                     & 1 & 2 & 3 & 4 & 5  \\ \hline 
      \hline

      \multirow{3}{*}{ISAL}                    & mAP                         &0.02366667 &0.12766667 &0.24633333 &0.32366667 &0.42 \\
                                               & bbox num                    &1338       &2777.66667 &3606.66667 &4446       &5616  \\ 
                                               & 10k $\times$ mAP / bbox num &0.17688092 &0.45961839 &0.68299445 &0.72799520 &0.74786325  \\ \hline

      \multirow{3}{*}{Coreset}                 & mAP                         &0.02366667 &0.13       &0.25733333 &0.38466667 &0.459 \\
                                               & bbox num                    &1338       &2624.66667 &4149.33333 &5736.66667 &7292.66667  \\
                                               & 10k $\times$ mAP / bbox num &0.17688092 &0.49530099 &0.62017995 &0.67054038 &0.62939940  \\ \hline

      \multirow{3}{*}{Random}                  & mAP                         &0.02366667 &0.11666667 &0.24133333 &0.342      &0.43666667 \\
                                               & bbox num                    &1338       &2683.66667 &4097.66667 &5450.66667 &6833.66667  \\ 
                                               & 10k $\times$ mAP / bbox num &0.17688092 &0.43472861 &0.58895306 &0.62744618 &0.63899322  \\ \hline

      \multirow{3}{*}{Learningloss}            & mAP                         &0.023      &0.13       &0.25233333 &0.35933333 &0.42433333 \\
                                               & bbox num                    &1338       &2780.66667 &4161.66667 &5627       &7236  \\ 
                                               & 10k $\times$ mAP / bbox num &0.17189836 &0.46751379 &0.60632759 &0.63858776 &0.58641975  \\ \hline     

      \multirow{3}{*}{Localization stability}  & mAP                         &0.02366667 &0.136      &0.243      &0.33233333 &0.4245 \\
                                               & bbox num                    &1338       &2713.33333 &3940.66667 &5653       &7601.66667 \\ 
                                               & 10k $\times$ mAP / bbox num &0.17688092 &0.50122850 &0.61664693 &0.58788844 &0.55843017  \\  
                                               
                                               \hline  \hline 
      \multicolumn{2}{|c|}{\multirow{2}*{Method}} & \multicolumn{5}{c|}{3 times average of results in each step} \\ \cline{3-7} 
      \multicolumn{2}{|c|}{~}                     & 6 & 7 & 8 & 9 & 10  \\ \hline 
      \hline

      \multirow{3}{*}{ISAL}                     & mAP                         &0.47166667 &0.515      &0.55233333 &0.57666667 &0.596 \\
                                                & bbox num                    &7190.33333 &8703.66667 &10160      &11503      &12967.6667  \\ 
                                                & 10k $\times$ mAP / bbox num &0.65597330 &0.59170465 &0.54363517 &0.50131850 &0.45960466  \\ \hline

      \multirow{3}{*}{Coreset}                  & mAP                         &0.51033333 &0.552      &0.57666667 &0.59666667 &0.604 \\
                                                & bbox num                    &8812       &10286.3333 &11635      &12888.3333 &14194.3333  \\
                                                & 10k $\times$ mAP / bbox num &0.57913451 &0.53663437 &0.49563100 &0.46295099 &0.42552192  \\ \hline

      \multirow{3}{*}{Random}                   & mAP                         &0.4845     &0.53533333 &0.56133333 &0.57866667 &0.595 \\
                                                & bbox num                    &8188.33333 &9575.66667 &11011.6667 &12426      &13813  \\ 
                                                & 10k $\times$ mAP / bbox num &0.59169550 &0.55905594 &0.50976237 &0.46569022 &0.43075364  \\ \hline

      \multirow{3}{*}{Learningloss}             & mAP                         &0.49566667 &0.54866667 &0.566      &0.58266667 &0.59866667 \\
                                                & bbox num                    &8752.33333 &10368      &11750      &12981.6667 &14119.3333  \\ 
                                                & 10k $\times$ mAP / bbox num &0.56632517 &0.52919239 &0.48170213 &0.44883811 &0.42400491  \\ \hline     

      \multirow{3}{*}{Localization stability}   & mAP                         &0.46533333 &0.52766667 &0.55933333 &0.59       &0.60466667 \\
                                                & bbox num                    &9169.66667 &11056.3333 &12580      &13853.6667 &14843.6667  \\ 
                                                & 10k $\times$ mAP / bbox num &0.50747028 &0.47725286 &0.44462109 &0.42588003 &0.40735667  \\ \hline                                                 

   \end{tabular}
    \centering
   \vspace{-3mm}
   \caption{The experiment results on VOC2012 with FCOS.}
   \vspace{-3mm}  
   \label{experiment:voc2012}   
\end{table*}

 \begin{table*}[!t]
   \centering
   \begin{tabular}{|c|c|ccccc|}
     \hline 
     \multicolumn{2}{|c|}{\multirow{2}*{Method}} & \multicolumn{5}{c|}{3 times average of results in each step} \\ \cline{3-7} 
     \multicolumn{2}{|c|}{~}                     & 1 & 2 & 3 & 4 & 5  \\ \hline 
      \hline

      \multirow{3}{*}{ISAL}                    & AP                         &0.12833333 &0.14433333 &0.153      &0.16233333 &0.166 \\
                                               & bbox num                   &36603.6667 &37934      &40028.3333 &42021      &43947.3333  \\ 
                                               & 10k $\times$ AP / bbox num &0.03506024 &0.03804854 &0.03822293 &0.03863148 &0.03777249  \\ \hline

      \multirow{3}{*}{Coreset}                 & AP                         &0.12833333 &0.15266667 &0.17333333 &0.18766667 &0.19833333 \\
                                               & bbox num                   &36603.6667 &47194      &57032.3333 &66384.3333 &75564.3333  \\
                                               & 10k $\times$ AP / bbox num &0.03506024 &0.03234875 &0.03039212 &0.02826972 &0.02624695  \\ \hline

      \multirow{3}{*}{Random}                  & AP                         &0.12833333 &0.15033333 &0.16566667 &0.17733333 &0.188 \\
                                               & bbox num                   &36603.6667 &44055.6667 &51381.3333 &58653.6667 &66240  \\ 
                                               & 10k $\times$ AP / bbox num &0.03506024 &0.03412350 &0.03224258 &0.03023397 &0.02838164  \\ \hline

      \multirow{3}{*}{Learningloss}            & AP                         &0.127      &0.147      &0.163      &0.17766667 &0.185 \\
                                               & bbox num                   &36603.6667 &62127.6667 &84566.3333 &105900.333 &126722.333  \\ 
                                               & 10k $\times$ AP / bbox num &0.03469598 &0.02366096 &0.01927481 &0.01677678 &0.01459885  \\ \hline     

      \multirow{3}{*}{Localization stability}  & AP                         &0.12833333 &0.149      &0.16566667 &0.179      &0.191 \\
                                               & bbox num                   &36603.6667 &47503.3333 &58252.6667 &69085.6667 &79574.6667  \\ 
                                               & 10k $\times$ AP / bbox num &0.03506024 &0.03136622 &0.02843933 &0.02590986 &0.02400261  \\

      \hline \hline 
      \multicolumn{2}{|c|}{\multirow{2}*{Method}} & \multicolumn{5}{c|}{3 times average of results in each step} \\ \cline{3-7} 
      \multicolumn{2}{|c|}{~}                     & 6 & 7 & 8 & 9 & 10  \\ \hline 
      \hline

      \multirow{3}{*}{ISAL}                    & AP                          &0.172      &0.18666667 &0.18333333 &0.189      &0.19133333 \\
                                                & bbox num                   &45810      &47864.6667 &49865      &51930.6667 &54414.3333 \\ 
                                                & 10k $\times$ AP / bbox num &0.03754639 &0.03899884 &0.03676594 &0.03639468 &0.03516230  \\ \hline

      \multirow{3}{*}{Coreset}                 & AP                          &0.20933333 &0.21766667 &N/A&N/A&N/A \\
                                                & bbox num                   &84924      &94075.6667 &N/A&N/A&N/A  \\
                                                & 10k $\times$ AP / bbox num &0.02464949 &0.02313740 &N/A&N/A&N/A  \\ \hline

      \multirow{3}{*}{Random}                  & AP                          &0.199      &0.20533333 &0.21433333 &0.22       &N/A \\
                                                & bbox num                   &73457      &80720.6667 &88030.6667 &95464.3333 &N/A  \\ 
                                                & 10k $\times$ AP / bbox num &0.02709068 &0.02543752 &0.02434758 &0.02304526 &N/A  \\ \hline

      \multirow{3}{*}{Learningloss}            & AP                          &0.19433333 &0.20233333 &0.21166667 &0.21766667 &N/A \\
                                                & bbox num                   &145197     &163798     &181040.333 &197804     &N/A  \\ 
                                                & 10k $\times$ AP / bbox num &0.01338412 &0.01235261 &0.01169169 &0.01100416 &N/A  \\ \hline     

      \multirow{3}{*}{Localization stability}  & AP                          &0.2        &0.20866667 &0.217      &N/A&N/A \\
                                                & bbox num                   &89557.3333 &99480.6667 &109179.333 &N/A&N/A  \\ 
                                                & 10k $\times$ AP / bbox num &0.02233206 &0.0209756  &0.01987556 &N/A&N/A  \\

                                                \hline \hline 
      \multicolumn{2}{|c|}{\multirow{2}*{Method}} & \multicolumn{5}{c|}{3 times average of results in each step} \\ \cline{3-7} 
      \multicolumn{2}{|c|}{~}                     & 11 & 12 & 13 & 14 & 15  \\ \hline 
         \hline
         \multirow{3}{*}{ISAL}                    & AP                       &0.19466667 &0.197      &0.20233333 &0.207      &0.20933333 \\
                                                & bbox num                   &56457      &59059      &61677      &64093.3333 &66729  \\ 
                                                & 10k $\times$ AP / bbox num &0.03448052 &0.03335647 &0.03280531 &0.03229665 &0.03137067  \\ 

                                                \hline \hline 
      \multicolumn{2}{|c|}{\multirow{2}*{Method}} & \multicolumn{5}{c|}{3 times average of results in each step} \\ \cline{3-7} 
      \multicolumn{2}{|c|}{~}                     & 16 & 17 & 18 & 19 & 20  \\ \hline 
         \hline
         \multirow{3}{*}{ISAL}                    & AP                       &0.21       &0.21133333 &0.216      &0.21633333 &0.218 \\
                                                & bbox num                   &69402      &72282.3333 &74690      &77545.6667 &80139  \\ 
                                                & 10k $\times$ AP / bbox num &0.03025849 &0.02923720 &0.02891953 &0.0278975  &0.02720274  \\ \hline

   \end{tabular}

   \vspace{-3mm}
   \caption{The experiment results on COCO with FCOS.}
   \vspace{-3mm}  
   \label{experiment:coco}   
\end{table*}

\begin{table*}[!t]
   \centering
   \begin{tabular}{|c|c|ccccc|}
     \hline 
     \multicolumn{2}{|c|}{\multirow{2}*{Method}} & \multicolumn{5}{c|}{Results in each step} \\ \cline{3-7} 
     \multicolumn{2}{|c|}{~}                     & 1 & 2 & 3 & 4 & 5  \\ \hline 
      \hline

      \multirow{3}{*}{ISAL}                    & AP                         &0.212      &0.25       &0.275      &0.291      &0.305 \\
                                               & bbox num                   &86838      &118762     &169688     &232923     &286589  \\ 
                                               & 10k $\times$ AP / bbox num &0.02441328 &0.02105050 &0.01620621 &0.01249340 &0.01064242  \\ \hline

      \multirow{3}{*}{Coreset}                 & AP                         &0.212      &0.273      &0.305      &0.321      &0.334 \\
                                               & bbox num                   &86838      &201072     &307552     &413095     &510927  \\
                                               & 10k $\times$ AP / bbox num &0.02441328 &0.01357723 &0.00991702 &0.00777061 &0.00653714  \\ \hline

      \multirow{3}{*}{Random}                  & AP                         &0.212      &0.264      &0.294      &0.309      &0.322 \\
                                               & bbox num                   &86838      &173507     &259539     &345013     &430922  \\ 
                                               & 10k $\times$ AP / bbox num &0.02441328 &0.01521552 &0.01132778 &0.00895618 &0.00747235  \\ \hline

      \multirow{3}{*}{Learningloss}            & AP                         &0.212      &0.271      &0.3        &0.319      &0.33 \\
                                               & bbox num                   &86838      &291934     &426039     &532231     &609475  \\ 
                                               & 10k $\times$ AP / bbox num &0.02441328 &0.00928292 &0.00704161 &0.00599364 &0.00541450  \\ \hline     

      \multirow{3}{*}{Localization stability}  & AP                         &0.212      &0.271      &0.296      &0.314      &0.327 \\
                                               & bbox num                   &86838      &194677     &289580     &385590     &485663 \\ 
                                               & 10k $\times$ AP / bbox num &0.02441328 &0.01392049 &0.0102217  &0.00814337 &0.00673306  \\       
                                               
                                               \hline \hline 
      \multicolumn{2}{|c|}{\multirow{2}*{Method}} & \multicolumn{5}{c|}{Results in each step} \\ \cline{3-7} 
      \multicolumn{2}{|c|}{~}                     & 6 & 7 & 8 & 9 & 10  \\ \hline 
      \hline

      \multirow{3}{*}{ISAL}                    & AP                          &0.322      &0.331      &0.347      &0.354      &0.363 \\
                                                & bbox num                   &351747     &449780     &579039     &719234     &860001  \\ 
                                                & 10k $\times$ AP / bbox num &0.00915431 &0.00735915 &0.00599269 &0.00492190 &0.00422093  \\ \hline

      \multirow{3}{*}{Coreset}                 & AP                          &0.344      &0.351      &0.355      &0.36       &0.364 \\
                                                & bbox num                   &601362     &680853     &748513     &806205     &860001  \\
                                                & 10k $\times$ AP / bbox num &0.00572035 &0.00515530 &0.00474274 &0.00446537 &0.00423255  \\ \hline

      \multirow{3}{*}{Random}                   & AP                         &0.332      &0.343      &0.349      &0.356      &0.362 \\
                                                & bbox num                   &516689     &602084     &688451     &774142     &860001  \\ 
                                                & 10k $\times$ AP / bbox num &0.00642553 &0.0056969  &0.00506935 &0.0045986  &0.00420930  \\ \hline

      \multirow{3}{*}{Learningloss}            & AP                          &0.338      &0.35       &0.35       &0.358      &0.361 \\
                                                & bbox num                   &668558     &713657     &751218     &805063     &860001  \\ 
                                                & 10k $\times$ AP / bbox num &0.00505566 &0.00490432 &0.0046591  &0.00444686 &0.00419767  \\ \hline     

      \multirow{3}{*}{Localization stability}  & AP                          &0.339      &0.344      &0.348      &0.358      &0.36 \\
                                                & bbox num                   &583831     &674087     &744716     &801876     &860001  \\ 
                                                & 10k $\times$ AP / bbox num &0.00580648 &0.00510320 &0.00467292 &0.00446453 &0.00418604  \\ \hline                                                 

   \end{tabular}
   \centering
   \vspace{-3mm}
   \caption{The experiment results on COCO in large-scale setting with FCOS.}
   \vspace{-3mm}  
   \label{experiment:cocofull}   
\end{table*}

\begin{table*}[!t]
   \centering
   \begin{tabular}{|c|c|ccccc|}
     \hline 
     \multicolumn{2}{|c|}{\multirow{2}*{Method}} & \multicolumn{5}{c|}{3 times average of results in each step} \\ \cline{3-7} 
     \multicolumn{2}{|c|}{~}                     & 1 & 2 & 3 & 4 & 5  \\ \hline 
      \hline
      \multirow{3}{*}{ISAL}                    & AP                         &0.17233333 &0.18266667 &0.19066667 &0.19766667 &0.203 \\
                                               & bbox num                   &36603.6667 &40120.3333 &43547.3333 &46840.6667 &50633.3333  \\ 
                                               & 10k $\times$ AP / bbox num &0.04708089 &0.04552970 &0.04378378 &0.04219980 &0.04009217 \\ \hline

      \multirow{3}{*}{Random}                  & AP                         &0.17233333 &0.18633333 &0.19866667 &0.207      &0.21533333 \\
                                               & bbox num                   &36603.6667 &44055.6667 &51381.3333 &58653.6667 &66240  \\ 
                                               & 10k $\times$ AP / bbox num &0.04708089 &0.04229498 &0.03866514 &0.03529191 &0.03250805  \\ \hline

      \multirow{3}{*}{Localization stability}  & AP                         &0.17333333 &0.18266667 &0.19133333 &0.20033333 &0.20633333 \\
                                               & bbox num                   &36603.6667 &42171      &47576.3333 &53144.6667 &59066.3333 \\ 
                                               & 10k $\times$ AP / bbox num &0.04735410 &0.04331571 &0.04021607 &0.03769585 &0.03493248  \\   

                                               \hline \hline 
      \multicolumn{2}{|c|}{\multirow{2}*{Method}} & \multicolumn{5}{c|}{3 times average of results in each step} \\ \cline{3-7} 
      \multicolumn{2}{|c|}{~}                     & 1 & 2 & 3 & 4 & 5  \\ \hline 
      \hline
      \multirow{3}{*}{ISAL}                    & AP                          &0.20733333 &0.21266667 &0.21766667 &N/A&N/A \\
                                                & bbox num                   &54060.3333 &57655      &61360.3333 &N/A&N/A  \\ 
                                                & 10k $\times$ AP / bbox num &0.03835221 &0.03688608 &0.03547351 &N/A&N/A  \\ \hline

      \multirow{3}{*}{Random}                  & AP                          &0.222      &N/A        &N/A&N/A&N/A \\
                                                & bbox num                   &73457      &N/A& N/A&N/A&N/A  \\ 
                                                & 10k $\times$ AP / bbox num &0.03022176 &N/A&N/A&N/A&N/A  \\ \hline

      \multirow{3}{*}{Localization stability}  & AP                          &0.21133333 &0.21766667 &N/A&N/A&N/A \\
                                                & bbox num                   &64631.3333 &70119.3333 &N/A&N/A&N/A \\ 
                                                & 10k $\times$ AP / bbox num &0.03269828 &0.03104232 &N/A&N/A&N/A  \\ \hline

   \end{tabular}

   \centering
   \vspace{-3mm}
   \caption{The experiment results on COCO with Faster R-CNN.}
   \vspace{-3mm}  
   \label{experiment:coco_fasterrcnn}   
\end{table*}

\end{document}